\documentclass[letterpaper, 10 pt, conference]{ieeeconf}  % Comment this line out if you need a4paper
\usepackage{booktabs}

\IEEEoverridecommandlockouts                              % This command is only needed if 
\usepackage{amsmath} % assumes amsmath package installed

\title{\LARGE \bf
Lipschitz Optimisation for Lipschitz Interpolation$^*$ 
}

\author{Jan-Peter Calliess$^{1}$ % <-this % stops a space
\thanks{*This paper is an extended version of a conference paper that will appear in the Proceedings of the American Control Conference (ACC 2017).}
\thanks{$^{1}$Jan-Peter Calliess is with the Engineering Department,
        University of Cambridge, UK.
        {\tt\small jpc73@cam.ac.uk}}%
}

\usepackage{subfigure}
\usepackage{amsmath}
\usepackage{amssymb}
\usepackage{amsthm}
\usepackage{tikz}
\newtheorem{thm}{Theorem}[section]

\newtheorem{lem}[thm]{Lemma}

\theoremstyle{definition}
\newtheorem{defn}[thm]{Definition}
\theoremstyle{remark}

\newcommand{\matnorm}[1]{{\left\vert\kern-0.25ex\left\vert\kern-0.25ex\left\vert #1 
    \right\vert\kern-0.25ex\right\vert\kern-0.25ex\right\vert}}
\newcommand{\opnorm}[1]{{\left\vert\kern-0.25ex\left\vert\kern-0.25ex\left\vert #1 
    \right\vert\kern-0.25ex\right\vert\kern-0.25ex\right\vert}}		

\newcommand{\norm}[1]{\left\Vert#1\right\Vert}

\newcommand{\abs}[1]{\left\vert#1\right\vert}

\newcommand{\Real}{\mathbb R}

\newcommand{\nat}{\mathbb N}

\newcommand{\argmin}{\text{argmin}}

\renewcommand{\d}[1]{\text{ d}#1}

\newcommand{\data}{\ensuremath{ \mathcal D} }

\newcommand{\dataeval}{\ensuremath{ \mathcal D^{\text{eval}}} }
\newcommand{\datacond}{\ensuremath{ \mathcal D^{\text{cond}}} }
\newcommand{\datatex}{\ensuremath{ \mathcal D^{\text{tex}}} }

\newcommand{\param}{\ensuremath{\theta}}%parameter for lin. Dif OP
\newcommand{\paramspace}{\ensuremath{\Theta}}%parameter for lin. Dif OP

\newcommand{\inspace}{\ensuremath{ \mathcal X}}
\newcommand{\outspace}{\ensuremath{ \mathcal Y}}

\newcommand{\indset}{\ensuremath{ \mathcal I}}
\newcommand{\indsett}{\ensuremath{ {\mathcal I_{t}}}}

\newcommand{\metric}{\, \mathfrak{d}} % distance metric
\newcommand{\predf}{\, \mathfrak{  \hat f}} % hypothesis
\newcommand{\predfn}{\, \mathfrak{  \hat f_n}} % hypothesis
\newcommand{\lossfct}{\, {\mathfrak l}} % loss function

\newcommand{\hestthresh}{\ensuremath{ \lambda}}
\newcommand{\domeval}{\, \mathcal X^{\text{eval}}} % evaluation domain
\newcommand{\domtest}{\, \mathcal X^{\text{test}}} %

\newcommand{\raneval}{\, \mathcal Y^{\text{eval}}} % evaluation range
\newcommand{\obserr}{\mathfrak e} % upper bound function on observational noise
\newcommand{\obserrbnd}{\bar{\mathfrak e}}

\newcommand{\decke}{\ensuremath{\mathfrak u}}
\newcommand{\boden}{\ensuremath{\mathfrak l}}

\renewcommand{\d}{\ensuremath{\text{ d}}}

\newcommand{\beq}{\begin{equation}}
\newcommand{\eeq}{\end{equation}}

\begin{document}

\maketitle
\thispagestyle{empty}
\pagestyle{empty}
\begin{abstract}
Techniques known as \emph{Nonlinear Set Membership} prediction, \emph{Kinky Inference} or \emph{Lipschitz Interpolation} are fast and numerically robust approaches to nonparametric machine learning that have been proposed to be utilised in the context of system identification and learning-based control. 
They 
utilise \emph{presupposed} Lipschitz properties in order to compute inferences over unobserved function values. Unfortunately, most of these approaches rely on exact knowledge about the input space metric as well as about the Lipschitz constant. Furthermore, existing techniques to estimate the Lipschitz constants from the data are not robust to noise or seem to be ad-hoc and typically are decoupled from the ultimate learning and prediction task.
To overcome these limitations, we propose an approach for optimising parameters of the presupposed metrics by minimising validation set prediction errors. To avoid poor performance due to local minima, we propose to utilise Lipschitz properties of the optimisation objective to ensure global optimisation success. The resulting approach is a new flexible method for nonparametric black-box learning.
We provide experimental evidence of the competitiveness of our approach on artificial as well as on real data.
\end{abstract}

\section{Introduction}

Supervised machine learning methods are algorithms for inductive inference. On the basis of a sample, they construct (learn) a computable model of a data generating process that facilitates inference over the underlying ground truth function and aims to predict its function values at unobserved inputs. 
%
%If the inferences (i.e. predictions of function values) are utilised in a decision-making process whose outcome involves risk, information about the prediction uncertainty can be vital. For instance, a robot that has a poor model about its dynamics would have to keep greater distance to obstacles than one that has a good model. Having such applications in mind, we are especially interested in learning algorithms that allow for conservative inference. That is, the predictions come with uncertainty quantifications that never underestimate the true uncertainty. 
%

Among supervised learning methods, nonparametric algorithms tend to offer greater flexibility to learn rich 
function classes. 
Unfortunately, many classical techniques for nonparametric regression, such as the \emph{Nadaraya-Watson estimator} \cite{Watson1964,Nadaraya1964} or the \emph{LOESS} method, \cite{Cleveland1979} suffer from a practical limitation: their regression performance depends on the choice of hyperparameters. While in principle, it would be possible to tune these to the data (in manner similar in spirit to the one we propose in this work), to the best of our knowledge, currently there is little understanding on how to do so with a global optimiser that offers theoretical performance guarantees on the optimisation solution. This means that in practice, one is left to engineer these hyperparameters (or the settings of an optimiser) by manual tuning in order to ensure good performance on a particular learning problem. Of course, this stands in opposition to the motivation for utilising nonparametric learning, especially in system identification: which is to facilitate flexible and fully automated black-box learning that does not require manual intervention.

Perhaps one of the most popular nonparametric machine learning method is Bayesian inference with \textit{Gaussian processes (GPs)} \cite{GPbook:2006}. GPs offer a flexible and principled probabilistic method for nonparametric regression and have evolved into one of the chief work-horses for learning dynamic systems \cite{Deisenroth2009,Deisenroth2011,Deisenroth2015,Tuongmodellearningsurvey2011,deisenrothsurvey2013,McHutchonthesis2014} in the research communities related to artificial intelligence and, more recently, also in control. 
However, they suffer from several limitations, including scalability to large data sets, a lack of understanding of how to bound the closed-loop dynamics resulting from controlling on the basis of a GP state-space model and, quite similar to the other classical regression methods,  the question of how to choose a good prior and its hyper-parameters in a principled yet computable manner. To alleviate the last problem, it is common practice to tune hyper-parameters of a chosen (typically universal) kernel to explain the data via the marginal log-likelihood \cite{GPbook:2006}. While often successful on many data sets, the result can be highly sensitive to the choice of optimiser, initialisations, data sets and computational budget. Unfortunately, little theoretical understanding of the important interplay between these components in the resulting inference mechanism seems to exist.

In contrast to such Bayesian methods this work builds on nonparametric regression techniques  that harnesses Lipschitz regularity of the target function to provide bounds on the predictions of the target function at unobserved inputs. Applied to machine learning the basic idea is that Lipschitz continuity constrains the set of possible function values of a target function at a query input, dependent on the distance between the query and the previously observed training examples. A prediction is then made by choosing a function value in the middle of the set of possible function values. This idea, at least going back to the era of ``Russian mathematics'' \cite{Sukharev1978}, has been redeveloped and advanced under different headlines including \emph{Lipschitz Interpolation} \cite{Zabinsky2003,Beliakov2006}, \emph{Nonlinear Set Membership (NSM)} interpolation methods \cite{Milanese2004} and \emph{Kinky Inference (KI)} \cite{calliess2014_thesis}. The presupposed Lipschitz constant as well as the assumed input space metric are crucial hyper-parameter choices of these methods that can drastically affect the predictive behaviour of trained inference rule. While a variety of Lipschitz constant estimators are known, they are designed independently from the prediction task and tend to be sensitive to noise. As an alternative, \cite{Milanese2004} proposed a method where the Lipschitz constant is estimated from a parametric model that is fitted to the data (e.g. a linear model or a neural network).
However, it remains unclear in how far this Lipschitz constant will aide the predictive performance of the Lipschitz interpolation model.

In this work, we propose a different approach. Firstly, we consider the more general problem of optimising for parameters of a chosen pseudo-metric. (As we will see the Lipschitz constant determination problem can be cast as a special case of this).
We then determine these free parameters by minimising an empirical estimate of the prediction error directly of the KI rule. When purposefully stating the error as an $\ell_1$-loss, we can derive a Lipschitz constant for the optimisation objective. This allows us to employ Lipschitz optimisation. In contrast, to other hyperparameter tuning approaches utilised in nonparametric machine learning, this global optimisation approach offers bounds on the optimisation success and can avoid falling into suboptimal solutions.

The result of this merger of Lipschitz optimisation for parameter optimisation and KI (based on the newly found parameters) is a reliable and fast nonparametric machine learning approach which we will refer to as \emph{Parameter Optimised Kinky Inference (POKI)}. 

Apart from the application to the automated determination of the Lipschitz constant in nonlinear set membership methods, we discuss other settings where our POKI approach uncovers   periodicities or inherent low-dimensionality (automated relevance determination) to learn good predictive models from data.
 
%\subsection{Related Work}
%Notice, one choice of parameter of the metric could be a Lipschitz constant (relative another metric).
%As the work perhaps most closely related to ours, 
%Milanese and Novara \cite{Milanese2004} propose an approach where the Lipschitz constant estimate is based on the maximum partial derivative of a parametric model that is fitted to the data. Then, noise bound parameters are fitted to explain the remaining empirical $l_2$-error. However, the approach seems to have several shortcomings. For example, the 
%authors do not explain how to perform the required optimisation steps to fit the Lipschitz constant. This might be non-trivial for all but the most simple parametric models. 
%Furthermore, the outcome of the optimisation will be very sensitive to the choice of parametric model, on the optimisation method employed to fit the model (i.e. on the parameters found) and to the bounds on the input space. 
%Unfortunately, the authors do not discuss these issues. By contrast, if we employed our more general approach to the task of finding a Lipschitz constant to the data on which to base our predictions, our approach directly fits the parameters of the prediction model to the data rather then going the indirect way of fitting a simpler surrogate parametric model first. Furthermore, our Lipschitz optimisation approach will guarantee optimal minimisation of the empirical prediction error -- something the other previous methods cannot.

\section{Kinky Inference and Lipschitz Interpolation}
\label{sec:KI_core}
In this section, we will rehearse the class of learning rules sometimes referred to as \emph{Kinky Inference}. They encompass a host of other methods such as Lipschitz Interpolation and Nonlinear Set Interpolation. 

\textbf{Setting.}
Let $f: \inspace \to \outspace$ be a \emph{target} or \emph{ground-truth} function we desire to learn and let $\inspace$, $\outspace$ be two spaces endowed with (pseudo-) metrics $\metric: \inspace^2 \to \Real_{\geq 0}, \metric_\outspace:\outspace^2 \to \Real_{\geq 0}$, respectively. 

For simplicity, we restrict our exposition to real-valued targets with $\metric_\outspace(y,y') = \abs{y-y'}$.

Assume that, at time step $n$, we have access to a \textit{sample} or \textit{data set} $\data_n:= \{\bigl( s_i, \tilde f_i \bigr) \, \vert \, i=1,\ldots, N_n \} $ containing $N_n \in \nat$ (possibly corrupted) sample values $\tilde f_i \in \outspace$ of \emph{target function} $f$ at sample input $s_i \in \inspace$. 
The sampled function values are allowed to have \textit{observational error} given by a (potentially stochastic) error function $\obserr : \inspace \to \Real^m_{\geq 0}$. That is, all we know is that $\tilde f_i = f(s_i)  + \obserr(s_i)$. For convenience, we may also write $\data_n = (\inspace_n,\outspace_n)$ where $\inspace_n = \{s_i | i=1,...,N_n\}\subset \inspace$ is the collection of sample inputs and $\outspace_n = \{\tilde f_i | i=1,...,N_n\} \subset \outspace$ is the sequence of observed function values.
It is our aim to learn target function $f$ in the sense that utilise the available data $\data_n$ to infer \textit{predictions} $\predfn(x)$ of $f(x)$ at unobserved \textit{query inputs} $x \notin \inspace_n$. In our context, the evaluation of $\predf_n$ is what we refer to as \textit{(inductive) inference} or \emph{prediction}. 

%
%For a discussion of the competing definitions of non-deductive inference in a philosophical context, the reader is referred to \cite{Flach2000}. 
%
The entire function $\predfn$ that is learned to facilitate predictions is referred to as the \textit{predictor}.

To ground the inference, a priori assumptions are necessary. In our context, we will generally assume that the target can be arbitrarily well approximated by Lipschitz continuous function with some Lipschitz constant. Remember, relative to our chosen pseudo-metrics, a real-valued function $\phi$ is Lipschitz continuous (with Lipschitz constant $\ell \geq 0$)  on domain $I \subset \inspace$ if $\metric_\outspace(\phi(x),\phi(x')) \leq \ell \metric(x,x'),\forall x,x'\in I$.
Note the metrics, may depend on a parameter $\param$. In this case, we highlight this dependency explicitly by writing $\metric(\cdot,\cdot;\param)$ instead of $\metric(\cdot,\cdot)$. Of course, the Lipschitz parameter can be absorbed into the parameter of a chosen pseudo-metric.

The approximation success of $\predfn$ relative to a target can be measure by various metrics. 
In this work, we will be most interested in the $\mathcal L_1$-\emph{prediction error}
$\mathcal E_1(\predfn;f)  := \norm{\predfn-f}_1 = \int_\inspace \abs{\predfn(x) - f(x)} \d x $. In practice, this error can be estimated by the \emph{ empirical prediction error} estimate 
\begin{equation}
\hat {\mathcal E_1}(\predfn;f) :=  \frac 1 {\abs{\inspace_{sample}}} \sum_{x \in \inspace_{sample}} \abs{\predfn(x) - f(x)} \label{eq:prederrest}
\end{equation} where the \emph{sample $\inspace_{sample}$} is a finite set of sample inputs. 

The average test set prediction error serves as a surrogate measure for the true prediction error; if the test set is chosen sufficiently dense (and the predictor and target are continuous) then $\hat {\mathcal E_1} \approx \mathcal E_1$. In the case of i.i.d. samples, we can also construe Eq. \ref{eq:prederrest} as a Monte-Carlo estimate of the $\mathcal L_1$- error.

In case we do not have access to a the target (or a noise-free sample), we might 
have to base our assessment of the method on the \emph{empirical sample prediction error }
\begin{equation}
\tilde {\mathcal E_1}(\predfn;\tilde f) :=  \frac 1 {\abs{\inspace_{sample}}} \sum_{x \in \domtest} \abs{\predfn(x) - \tilde f(x)}.  \label{eq:sampleprederr}
\end{equation}

\textbf{Learning rule.}
In this work we will expand on the basis of a special case of the class of kinky inference predictors \cite{calliess2014_thesis} to perform learning as inference over unobserved function values. The prediction rule can be stated as follows: 

\begin{defn}[Kinky inference (KI) rule (simplified) ] \label{def:KILsimplified}
Given access to a sample set $\data_n$ and an input space pseudo-metric $\metric(\cdot,\cdot;\param(n)): \inspace^2 \to \Real$ parameterised by $\param(n)$, we define the KI predictor by $\predfn \bigl(\cdot;\param(n),\data_n\bigr): \inspace \to \outspace$ to perform inference over function values as per:
	\begin{eqnarray*}
   \predfn\bigl(x; \param(n),\data_n\bigr) := \frac{1}{2} \decke_n(x;\param(n)\bigr) + \frac{1}{2}  \boden_n(x;\param(n)\bigr). \label{eq:KIpred_basic}
	\end{eqnarray*}
	Here, $\decke_n\bigl(\cdot;\param(n) \bigr), \boden_n\bigl(\cdot;\param(n) \bigr): \inspace \to \Real^m$ are called ceiling and floor functions, respectively. They are given by
	$\decke_{n}\bigl(x; \param(n)\bigr) := \min_{i=1,\ldots,N_n}   \tilde f_{i} +  \metric(x,s_i; \param(n)) $ and 
	$\boden_{n}\bigl(x; \param(n)\bigr) := \max_{i=1,\ldots,N_n}   \tilde f_{i} - \metric(x,s_i; \param(n)) $, respectively.  
 \end{defn}

In the literature, various generalisations and special cases exist. For instance, Calliess \cite{calliess2014_thesis} proposes a generalised framework called \emph{Kinky Inference}, that also allows for the specification of additional parameters. These include functions that allow the incorporation of a priori knowledge about upper and lower bounds on the target as well as upper and lower bounds on observational noise. 

A special case arises for the choice of $\metric(x,y;\param(n) ) = \ell(n) \norm{x-y} $ which is referred to as \emph{Lipschitz Interpolation} \cite{Beliakov2006} or as \emph{Nonlinear Set Interpolation} \cite{Milanese2004}. Here the parameter $\param(n) =\ell(n)$ is the supposed Lipschitz constant of the target. Typically this constant is assumed to be either known a priori or estimated lazily from the data, e.g. ~\cite{Strongin1973} as follows:  \begin{equation}\label{eq:lazyconstadaptdefnonoise}
\hat \ell(n) := \max_{i \neq j} \frac{\abs{\tilde f_i -\tilde f_j}}{\norm{s_i-s_j}} .
\end{equation} 
Unfortunately, the latter estimate has the problem of being unbounded in the presence of observational noise. In the case of bounded noise, \cite{calliess2014_thesis} proposed to utilise the alternative estimator $\ell(n) := \max_{i \neq j} \frac{\abs{\tilde f_i -\tilde f_j} -2\obserrbnd}{\norm{s_i-s_j}}  $ where $\obserrbnd := \sup_x \abs{\obserr(x)}$ is an upper bound on the (zero-mean) noise. While this prevents the estimates to blow up in the bounded noise case, generally it is not clear how to obtain $\obserrbnd$. And, if $\obserrbnd$ is chosen too conservatively large then the prediction quality is questionable.
As an alternative approach, we will employ Lipschitz optimisation to find a parameter $\param (n)$ that minimises empirical prediction error on a validation data set.
%To give an intuition of this interpolation rule, consider the following special case where we have access to a noise-free sample $\data_n$ and suppose 
%the target $f$ is a real-valued $L^*$- Lipschitz continuous function. Observing the noise-free sample point $(s_i,f_i)$ constrains the set of function values $f(x)$ to the set $\mathbb S_i(x) =\{ \phi \in \outspace | \metric_\outspace(\phi, f_i) \leq L^* \metric(s_i,x) \}$. Considering a set of sample points $\data_n$, target value $f(x)$ is constrained to lie in the intersection $\mathbb S(x)=\cap_{i=1}^{N_n} \mathbb S_i(x)$. It is easy to see that the floor and ceiling functions are tight lower and upper bounds of $\mathbb S(x)$ with $\mathbb S(x) := \{ \phi \in \outspace | \boden_n(x; L^*) \leq \phi \leq \decke_n(x;L^*)\}$. In other words, setting parameter $L(n)$ to the best H\"older constant $L^*$ and bounds $\lbf =-\infty,\ubf=+\infty$ yields a predictor $\predfn(x)$ that for every query $x$ chooses the mid-point of the set $\mathbb S(x)$ of those function values that can possibly be assumed by a H\"older continuous function that interpolates the observed sample. Prediction error $\prederrn(x)$ simply is the radius of the set.
\section{Parameter Optimised Kinky Inference (POKI)}
%Above we have introduced kinky inference (KI) as a nonparametric machine learning method. 
%In order to infer function values, several parameters had to be specified a priori. In particular, we had to specify a pseudo-metric $\metric_\inspace$ as well as its parameter $\param$.
%
%As we will illustrate below, these choices determine much of the predictive performance of the resulting inferences. 
%However, in many real-world situations, the exact value of parameter $\param$ may well be unknown.
%So, in the absence of accurate knowledge of the suitable parameter $\param$ of the pseudo-metric, it would be helpful to estimate it from the data. A standard approach commonly adopted in machine learning is to search for a parameter $\param$ that promises to optimise predictive performance quantified by a suitable loss function. 

Given some separate data sets $\datacond$ and $\dataeval$, we will aim to choose a parameter of the pseudo-metric that minimises the empiricial sample prediction  error.
That is, we choose our parameter to be
 \begin{equation} \label{eq:lossminimiserparam}
\hat \param := \argmin_{\param \in \paramspace} \lossfct(\param;\datacond,\dataeval)
\end{equation}
where $\paramspace$ denotes a predefined parameter space and the loss quantifies the empirical sample prediction error  quantified by the loss function $\lossfct(\cdot;\datacond,\dataeval):\inspace \to \Real$ with
\begin{equation}
 \lossfct(\param;\datacond,\dataeval) =  \frac 1 {\abs{\domeval}} \sum_{x \in \domeval} \abs{  \tilde f(x) - \predfn(x;\param, \datacond)}.
\end{equation}
Here, we refer to $\dataeval = \bigl(\domeval,\raneval\bigr)$ as the available \emph{parameter evaluation data} and $\datacond$ as the \emph{conditioning data}.

To avoid overfitting and exact interpolation of the noise, we will generally ensure that the conditioning and evaluation data are different, i.e. at least $\dataeval \neq \datacond$. Ideally, they should be drawn independently from each other. 
We will give an illustration of the utility of this approach below. For now, we will confine ourselves to describe the approach we propose:

\textbf{Def. (Parameter Optimised Kinky Inference (POKI)):}
Assume at trial $n$, we are given access to a set of training examples $\data_n$ of size $N_n$. 

We will undergo the following steps:

\begin{enumerate}
\item We randomly partition the data into two sub sets: a \emph{conditioning data} set $\datacond_n$ and a parameter \emph{evaluation data} set $\dataeval_n$, $\datatex_n = \datacond_n \, {\dot \cup} \,  \dataeval_n$. Unless explicitly stated otherwise, both sets will be made close to equal in size.
%That is, we usually ensure that $\abs{\datacond_n } \in \bigl[\abs{\dataeval_n} -1,\abs{\dataeval_n}+1\bigr]$.  
\item Utilising the conditioning and evaluation data sets, we compute the minimum-loss parameter estimate $\hat \param_n := \argmin_{\param \in \paramspace} \lossfct(\param;\datacond_n,\dataeval_n)$ (cf. Eq. \ref{eq:lossminimiserparam}).
\item For prediction of future query inputs $q \in \inspace$, we use the KI prediction $\predfn(q;\hat \param_n,\data_n)$ (cf. Def. \ref{def:KILsimplified}), utilising the learned newly identified parameter estimate $\hat \param_n$ and conditioning on the full set of available data.
\end{enumerate}

We refer to the resulting predictor $\predfn(\cdot;\hat \param_n,\datatex_n)$ as a \emph{Parameter Optimised Kinky Inference (POKI)} rule.

%
%\subsection{A probabilistic interpretation}
%Notice we can interpret the minimisation of the chosen loss function as the task of finding the maximum likelihood estimate of the parameter under a suitably chosen density. 
%To this end, assume the data was drawn independently from a Laplacian distribution with the mode coinciding with the posterior prediction. That is, a noisy function value observation $\tilde f(x) \in \Real$ at input $x $ is assumed to have density $p(\tilde f(x)| x,\param, \datacond) = \frac{1}{2\lambda}\exp(-\frac{\abs{\tilde f(x) - \predfn(x;\param,\datacond)}}{\lambda})$ for some length scale parameter $\lambda$.
%Given some set of \emph{parameter evaluation data} $\dataeval = \bigl\{ ( x_i, \tilde f(x_i) ) | \, i=1,\dots, N \bigr\}$ we could determine the estimate by maximising the conditional likelihood  
%$p\bigl(\tilde f(x_1),\ldots,\tilde f(x_N)  | x_1,...,x_N,\param,\datacond \bigr) = \frac{1}{2\lambda}\exp \Bigl(-\prod_{i=1}^N\frac{\abs{\tilde f(x_i) - \predfn(x_i;\param,\datacond)}}{\lambda} \Bigr) $.
%This can be done be minimising the negative log-likelihood. Since multiplication with positive factors does not change the minimiser, we thus obtain the maximum likelihood estimate as:  
%\begin{eqnarray}
%\argmin_\param -\frac{\lambda}{N}\log p\bigl(\tilde f(x_1),\ldots,\tilde f(x_N)  | x_1,...,x_N, \param,\datacond\bigr) \\= \argmin_\param \sum_{i=1}^N\frac{\abs{\tilde f(x_i) - \predfn(x_i;\param,\datacond)}}{N} = \hat \param
%\end{eqnarray}
%where $\hat \param$ is the loss minimiser defined in Eq. \ref{eq:lossminimiserparam}.
%

\subsection{Lipschitz optimisation and constant determination}
%The POKI approach requires finding a pseudo-metric parameter estimate as the minimiser of a loss function as per Eq. \ref{eq:lossminimiserparam}. Unfortunately, as we will see below, the loss function may well be quite ``spiky'' and contain numerous local minima. Therefore, the choice of optimiser might well be crucial to guarantee a good parameter estimate $\hat \param$.
In order to guarantee the quality of the parameter estimate $\hat \param_n$ even in the presence of non- differentiabilities and local optima of the loss function, we propose to utilise Lipschitz optimisation methods. In one dimensional settings we could for instance utilise Shubert's method \cite{Shubert:72}. 
Various extensions to the multi-dimensional case exist including stochastic methods that offer probabilistic bounds that allow for guarantees with computational effort scaling linearly with the number of dimensions \cite{Zhang1995} as well as recent local approaches based on regret analysis \cite{Munos2014}. In this work we use the simple approach sketched in Sec. 2.5 of  \cite{calliess2014_thesis}. While the effort to reduce the given worst-case bound scales exponentially in the number of dimensions, the approach has the advantage of being simple and giving deterministic bounds. However, we intend to try out the more advanced, scalable approaches to Lipschitz optimisation in the course of future work.

In our chosen Lipschitz optimisation method, if we know a Lipschitz constant $L(\Omega)$ of the objective function \[\Omega: \begin{cases} \paramspace \to \Real \\ \param \mapsto \lossfct(\param;\datacond_n,\dataeval_n) \end{cases}\] then we can find a guaranteed global minimum up to a worst-case error bound that can be pre-specified in advance, thereby avoiding the pitfall of local minima.

To this end, we need to determine the Lipschitz constant of $\Omega$, i.e. of the loss as a function of the parameter $\param$. We denote the Lipschitz constant of a function $f$ by $L(f)$. For simplicity, we choose the metric induced by the maximum-norm to define Lipschitz continuity of $\Omega$. Hence, for a given loss we desire to derive a nonnegative number $L(\Omega) \in \Real_+$ such that:
$\forall \param,\param' \in \paramspace: \abs{\Omega(\param) - \Omega(\param')} \leq L(\Omega) \,\norm{\param -\param'}_\infty.$

Let the conditioning data be given by $\datacond_n = \{ (\mathfrak x_i,\mathfrak y_i) | i=1,\ldots, M_n \}$ and the evaluation data be given by $\dataeval_n = \{ ( x_i, y_i) | i=1,\ldots, Q_n \}$ such that the full data is $\data_n = \{(s_i,\tilde f_i) | i=1,...,N_n\} = \datacond_n \cup \dataeval_n$.

Appealing to the definition of the loss and the rules stated in Lem. \ref{lem:Hoeldarithmetic} (refer to the appendix), we see that \begin{equation} \label{eq:LOmegaasmaxLvarphii}
L(\Omega) \leq \max\{L(\varphi_i) \,|\, i=1,\dots,Q_n\}  
\end{equation}
where 
 $\varphi_i: \param \mapsto \predfn(x_i;\param,\datacond_n)$
 is the prediction of the $i$th evaluation example input as a function of the parameter $\param$.
We will now bound the Lipschitz constant of each $\varphi_i$:

By definition of the predictor we have \begin{align}
\predfn(q;\param;\datacond_n) &= \frac 1 2 \bigl( \mathfrak y_j +  \metric(\mathfrak x_j,q;\param)   \bigr) + \frac 1 2 \bigl( \mathfrak y_k -  \metric(\mathfrak x_k,q;\param) \bigr)
\end{align}
for some indices $k,j \in\{1,\ldots,M_n\}$ and all query inputs $q \in \inspace$.

Appealing to Lem. \ref{lem:Hoeldarithmetic}.8), we thus see that that for all $i$ we have 
\begin{align}\label{eq:Lvarphiibnd2}
L(\varphi_i) &\leq \max \{L\bigl(\metric(x_i,\mathfrak x_k;\cdot)\bigr) \,| k = 1,\dots,M_n \} 
\end{align}
where each $\metric(x_i,\mathfrak x_k;\cdot)$ is a function of the parameter of the pseudo-metric $\metric$. Since the Lipschitz constant of the loss is bounded from above by the maximum of the Lipschitz constants of all the $\varphi_i$ (cf. Eq. \ref{eq:LOmegaasmaxLvarphii}) and since $\datacond_n,\dataeval_n \subset \data_n$, we obtain the bound: 
\begin{align}\label{eq:LOmegaofmetric}
L(\Omega) 
%&\leq \max \{L\bigl(\metric(x_i,\mathfrak x_k;\cdot)\bigr) \,| \,i=1,\dots,Q_n,k =1,\dots,M_n \}\\
&\leq \max \{L\bigl(\metric(s_j,s_k;\cdot)\bigr) \,| \,j,k =1,\dots,N_n \}
\end{align}
where, for every pair $s_j,s_k$ of inputs in the available data, $\metric(s_j,s_k;\cdot)\bigr)$ is a function of hyper-parameter $\param$.
So, 
the determination of a bound on the desired Lipschitz constant depends on the conditioning data and the arithmetic definition the parameter $\param$ within the chosen pseudo-metric $\metric$. 
For several cases of general interest, we will next provide some Lipschitz constant derivations. The best Lipschitz constant of a function $\phi$ will be denoted by $L(\phi)$.
\begin{figure*}
        \centering
%                  				  \subfigure[ ]{
%    \includegraphics[width = 4.5cm,height = 4.5cm]
%								{content/pics/peroptnoise_wrong_par.pdf}
%    \label{fig:peroptnoise_wrong_par}
%  } 	
    		  \subfigure[  ]{
    \includegraphics[ width = .3\textwidth]
								{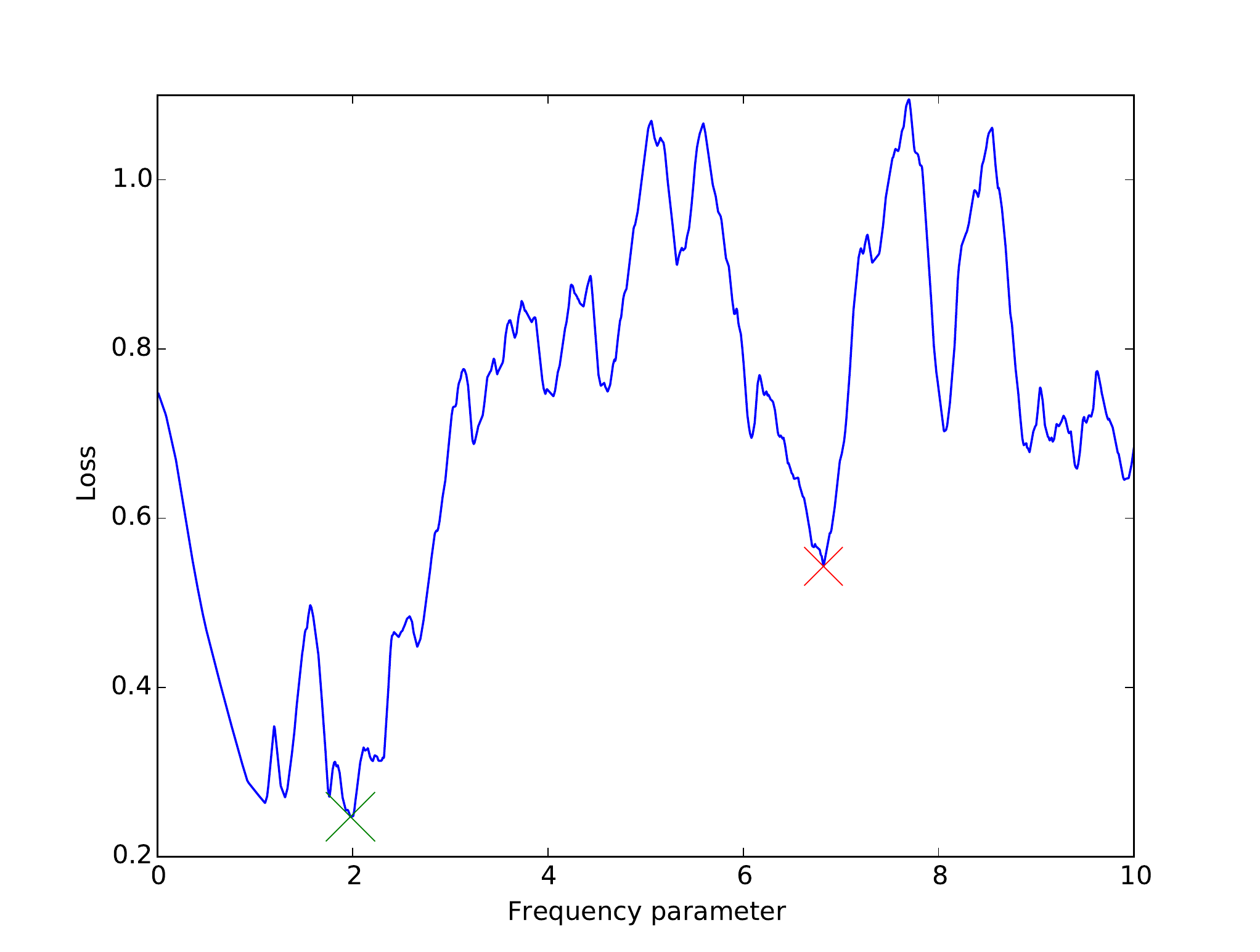}
    \label{fig:peroptnoise_loss}
  } 	
  		        				  \subfigure[ ]{
    \includegraphics[ width = .3\textwidth]
								{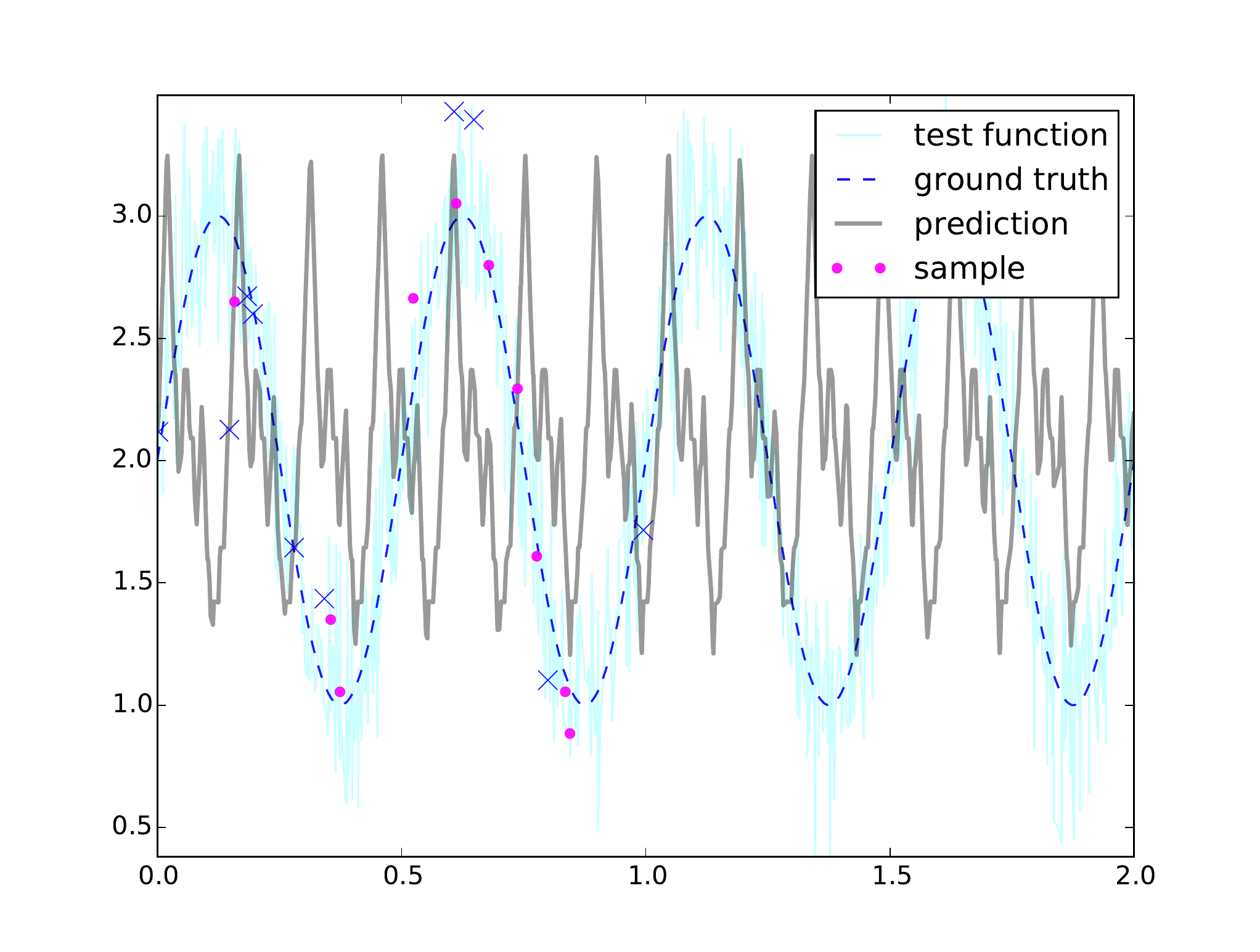}
    \label{fig:peroptnoise_POKI_Optimjl}
  } 	
  		  \subfigure[  ]{
    \includegraphics[ width = .3\textwidth]
								{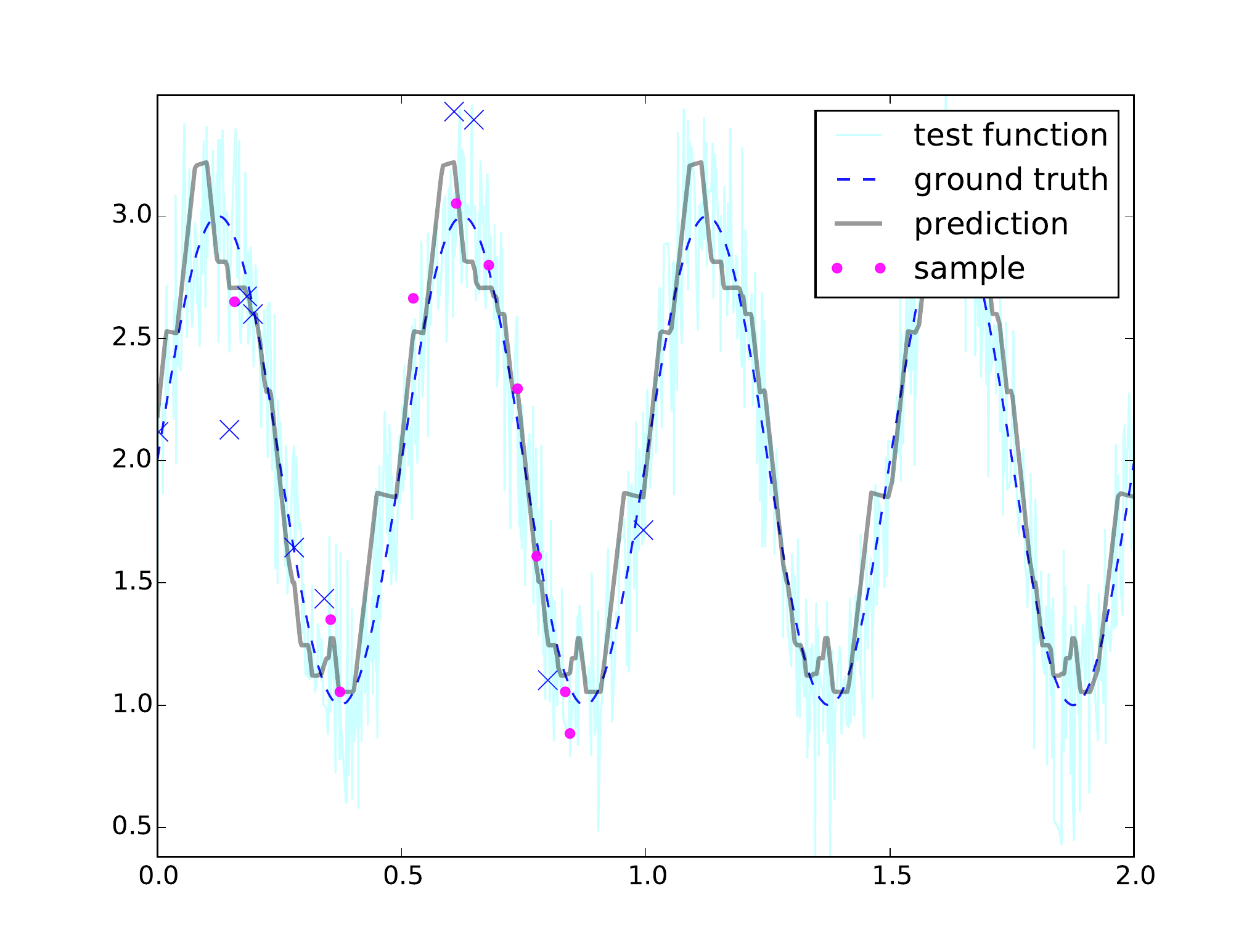}
    \label{fig:peroptnoise_POKI}
  } 	
   \caption{Comparison of POKI with Lipschitz optimisation v.s. a local optimiser to determine the frequency parameter based on a sample of 20 noisy data points of the periodic ground truth target function $f(x) = 2+ \sin(4\pi x)$ with frequency 2.
   Fig. \ref{fig:peroptnoise_loss} depicts the pertaining loss function (blue curve) as well as the minimising parameters found by the Lipschitz optimiser (green 'x') and Brent's method (red 'x'). Note that the Lipschitz optimiser managed to infer the true frequency as the global optimum of the loss. In contrast, Brent's method only found a local minimum giving rise to an overly large frequency.
   The other two plots show the corresponding 
   prediction results with POKI based the parameter found by Lipschitz optimisation (Fig. \ref{fig:peroptnoise_POKI}) and Brent's method (Fig. \ref{fig:peroptnoise_POKI_Optimjl}). The cyan plot is the noisy observational function $\tilde f$. The conditioning data is marked with 'x' while the parameter evaluation data is marked with magenta dots. Note, the usage of the parameter found by Lipschitz optimisation translates to a better fit of the predictor (grey line) to the target (dashed line).}
   \label{fig:peroptnoise}
\end{figure*}	 

\begin{itemize}
\item \textbf{Pseudo-metric for periodic functions:} If we know the target to be periodic with some frequency we may choose the pseudo-metric $\metric(x,x';\param) = \abs{\sin(\pi \param \abs{x-x'})} $. Knowing that taking the absolute value does not change a Lipschitz constant and that the Lipschitz constant of a differentiable function is the supremum of the absolute value of its derivative, we see that $L\bigl(\metric(x,x';\cdot)\bigr) \leq \pi \abs{x-x'} $. With reference to Eq. \ref{eq:LOmegaofmetric} it follows that in our case, a Lipschitz bound on the loss is given by $\pi$ times the diameter of the sample input set: 
$L(\Omega) \leq \pi \max \{\abs{s_j-s_k} \,| \,j,k =1,\dots,N_n \}.$

Note, the parameter $\param$ determines the frequency with which the pseudo-metric cannot distinguish two inputs in a periodic input space. 
Choosing a parameter $\param$ that matches the frequency of the target would be important to achieve adequate predictive performance.
Conversely, if one presupposes a false frequency $\param \neq 2$ then the prediction performance can suffer dramatically. Therefore, when optimising the parameter finding avoiding local minima is critical. Here, our global Lipschitz optimisation approach shines. For an illustration, refer to Fig. \ref{fig:peroptnoise}.
\item \textbf{Automated Lipschitz constant determination:}
As mentioned above, we might seek to perform Lipschitz interpolation with automated Lipschitz constant determination. To this end, we might choose the metric $\metric(x,y;\param_n ) = \param_n \norm{x-y}$ where parameter $\param$ acts as the assumed Lipschitz constant of the predictor. 
Clearly, we have $L(\metric(s_i,s_k;\cdot)) \leq \abs{s_j-s_k}$. Thus, the Lipschitz constant of the optimisation objective $\Omega$ can be bounded by the diameter of the sample input set $\inspace_n$: 
$L(\Omega) \leq  \max \{\abs{s_j-s_k} \,| \,j,k =1,\dots,N_n \} $.

We refer to the resulting POKI inference rule as \emph{POKI-LC} (where \emph{LC} stands for Lipschitz constant). Illustrations of the performance of this approach in contrast other methods are provided in the experimental section.
\item \textbf{Automated Relevance Determination (ARD).} A more general case of Lipschitz parameter determination is Automated Relevance Determination (ARD). Here, the goal is to find weights encoding the extent to which input space parameters are relevant for the prediction. This is interesting in high-dimensional input space where the input vectors might be features whose predictive role might be unclear a priori. To facilitate ARD, we might choose the metric $\metric(x,x';\param) = \max_{i=1,\ldots,d} \param_i \abs{x_i-x_i'}$. The weight $\param_i$ quantifies the degree to which the $i$th input component should contribute to the prediction. Large weights suggest a strong degree of importance: i.e. small deviations of a query $x_i$ from the closest example $s_{j,i}$ in the $i$th component will result in large uncertainty of the prediction. Conversely, if $\param_i=0$ effectively disables any influence of the $i$th input dimension in the prediction process.
 
Analogously to the kernel literature, inferring $\param$ from the data will be referred to as \emph{automated relevance determination (ARD)} and the resulting POKI rule will be referred to as \emph{POKI-ARD}.

To facillitate Lipschitz optimisation of the parameter vector, we need to derive a Lipschitz constant bound. Again appealing to the rules of Lipschitz arithmetic, we see that 
$L(\metric(s_i,s_k;\cdot)) \leq \norm{s_i-s_k}_\infty$ and thus,
$L(\Omega) \leq \max \bigl\{\norm{s_{j} - s_{k}}_{\infty} \,| \, j,k =1,\dots,N_n \bigr\}.$
In other words, the Lipschitz constant of the loss is bounded from above by the diameter of the data relative to the standard maximum-norm. 

%================ POKI-ARD EXAMPLE 1Pend ====================================

As a first illustration, consider learning the dynamics of a friction-less pendulum based on a sample. Denoting the angle of a pendulum by $q$,  its dynamics are given by the ordinary differential equation $\ddot q = f(x)$ where [m;b;l;g] 
$x = [q;\dot q]$ is called the state and 
$f(x) = - 9.81 \sin(x_1)$. Here, we learn the dynamics by learning the function $f$ based on a sample of triplets of angles, velocities and accelerations. We can see that this target function only depends on the first component of its input, namely the angle. So, if we incorporate this in our KI method by choosing the metric $\metric(x,x';\param) = \max\{\param_1 \abs{x_1-x_1'},\param_2\abs{x_2-x_2'} \}$ with $\param =[1; 0]$ we should expect superior predictive power over using the standard maximum norm (where $\param =[1; 1]$). 
Therefore, we expect ARD method to automatically uncover this advantageous parameter. As an 
example affirming this intuition, refer to Fig. \ref{fig:LACKIARD_1pend}.
The example of Fig. \ref{fig:LACKIARD_1pend_target} - Fig. \ref{fig:LACKIARD_1pend_pred1} illustrate how the ARD method can uncover the fact that the target function only depends on one of the input components and how this can affect the predictive performance. 
\begin{figure*}
        \centering
        				  \subfigure[Target function.]{
    \includegraphics[width = .3\textwidth,trim = 19mm 15mm 19mm 19mm, clip]
								{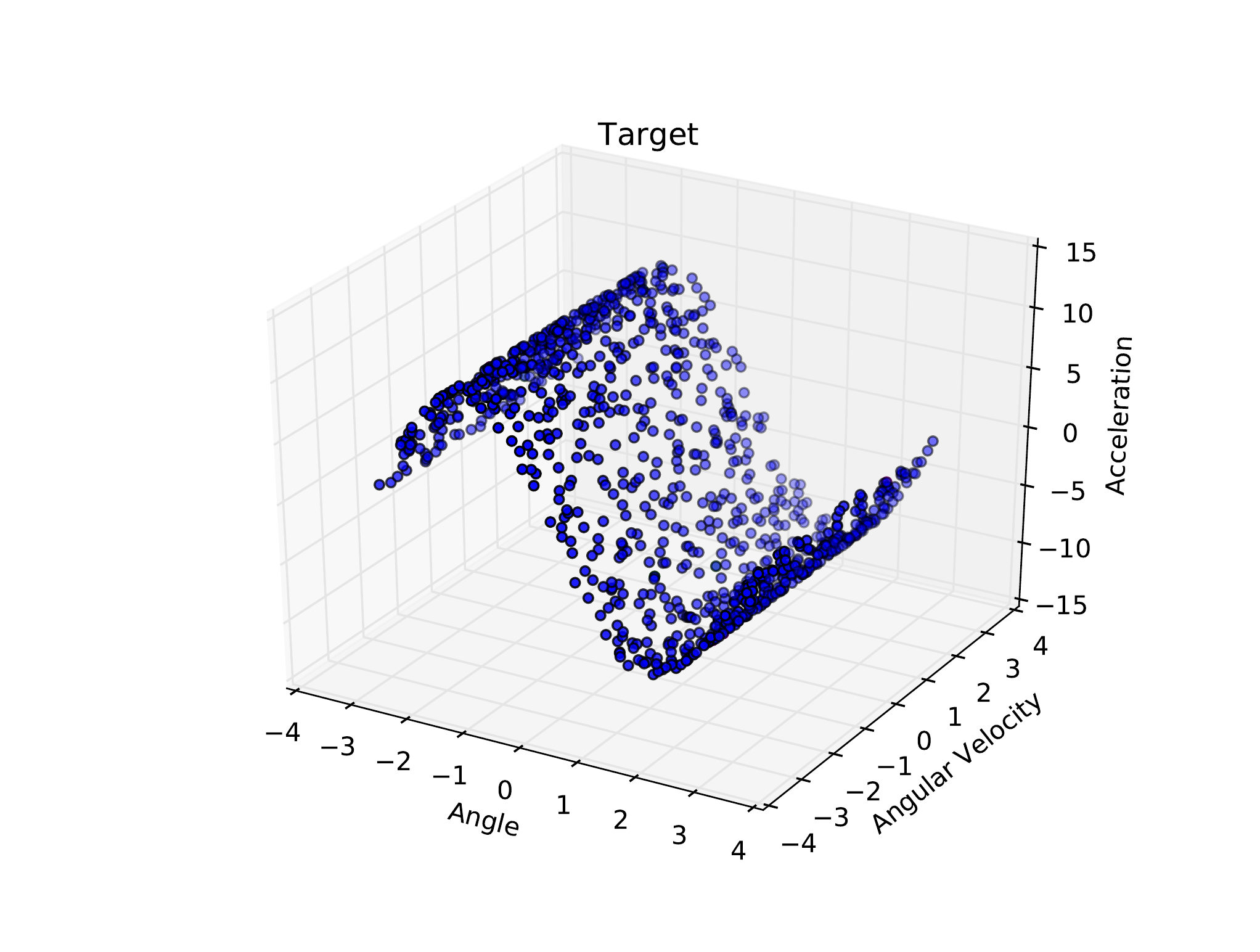}
    \label{fig:LACKIARD_1pend_target}
  } 	
  				  \subfigure[KI with estimated Lipschitz constant.]{
    \includegraphics[width = .3\textwidth,trim = 19mm 15mm 19mm 19mm, clip]
								{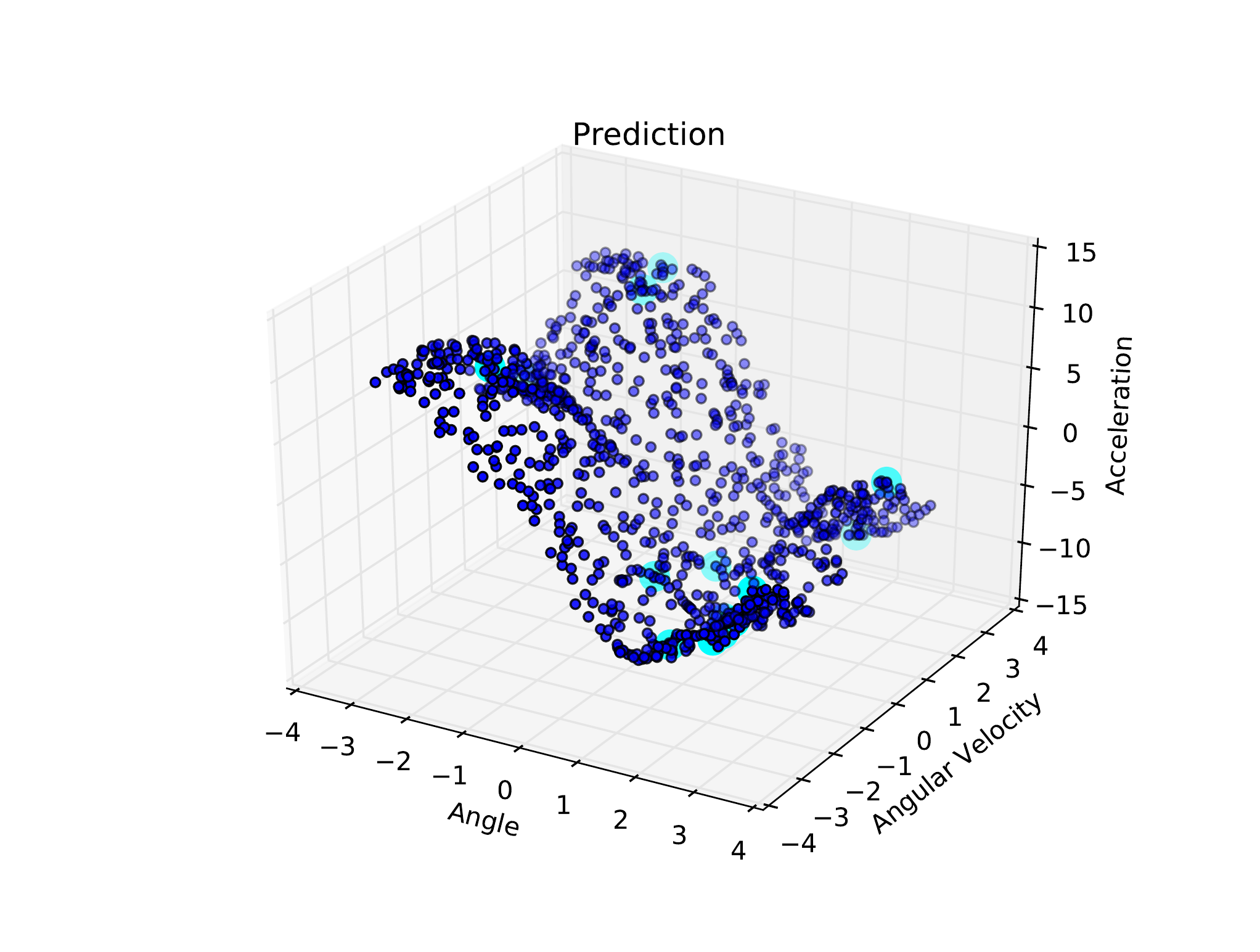}
    \label{fig:LACKIARD_1pend_pred0}
  } 
    				  \subfigure[POKI-ARD.]{
    \includegraphics[width = .3\textwidth,trim = 19mm 15mm 19mm 19mm, clip]
								{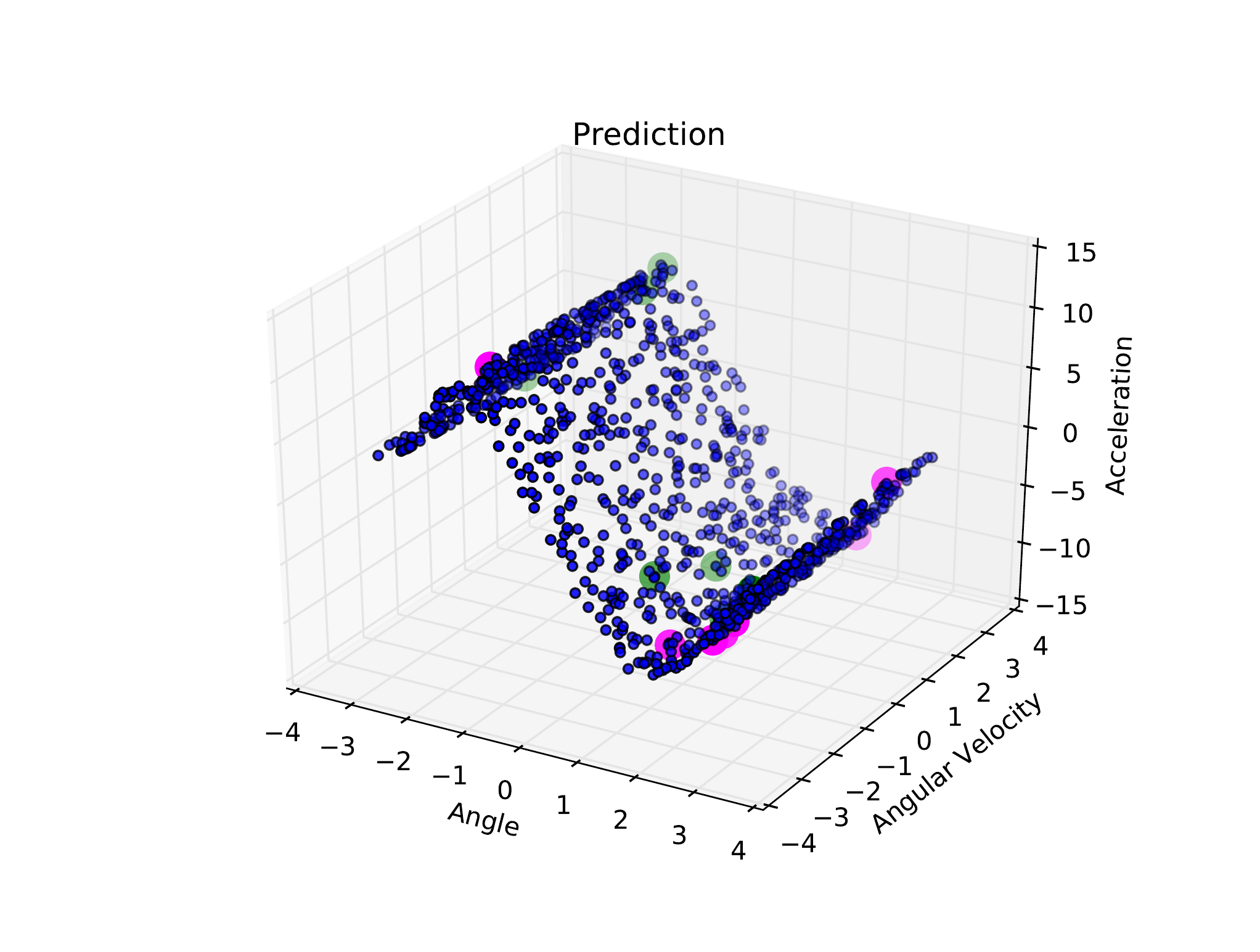}
    \label{fig:LACKIARD_1pend_pred1}
  } 
%      				  \subfigure[Comparison over 1000 rnd trials.]{
%    \includegraphics[width = 5.5cm]
%								{content/figs/ARD1pend_bp.pdf}
%    \label{fig:LACKIARD_1pend_bp}
%  } 

   \caption{LACKI with automated relevance determination.  Fig. \ref{fig:LACKIARD_1pend_target} depicts the target on a test set of M=1000 sample points (plotted in blue). Fig. \ref{fig:LACKIARD_1pend_pred0} depicts the prediction of kinky inference where the Lipschitz parameter was adjusted with Strongin's estimate. The sample contained 14 data points (cyan) of triples of angles, angular velocities and accelerations. Fig. \ref{fig:LACKIARD_1pend_pred1} depicts the pertaining prediction of POKI-ARD which determined the parameters to be $\param_1 = 0.99$ and $\param_2 = 0.02$. This accurately reflects the ground-truth model where accelerations are only dependent on the angle, not on the velocity.}\label{fig:LACKIARD_1pend}
\end{figure*}
%================ POKI-ARD EXAMPLE 1Pend ENDE ====================================

\end{itemize}

\section{Experiments}
In this section, we compare our approach to a number of well-established machine learning methods both on artificial and on real data.\\
\subsection{Artifical data.}

In order to have access to the ground-truth, we first tested the method on a sequence of artificial benchmark regression tasks. As the ground truth, we chose the target function $f: \inspace :=[0,1]^d \to \Real, x \mapsto \abs{-\cos(2\pi \, x_1)}+x_1$. Its values have a linear trend as well as a strongly nonlinear one, varying only with the first input component.

The data was obtained by artificially sampling uniformly $\inspace$ and superimposing the function values with i.i.d. zero-mean Gaussian noise with standard deviation $\frac 1 4$.

That is the sample was obtained from randomly selected inputs of the noisy data-generating  ``test function''  
$\tilde f(x) = f(x) + \nu(x)$ where $\nu(x) \stackrel{i.i.d.} \sim \mathcal N(0,\frac 1 {16})$.

\begin{figure}[h]
\centering
\includegraphics[width = .32\textwidth]{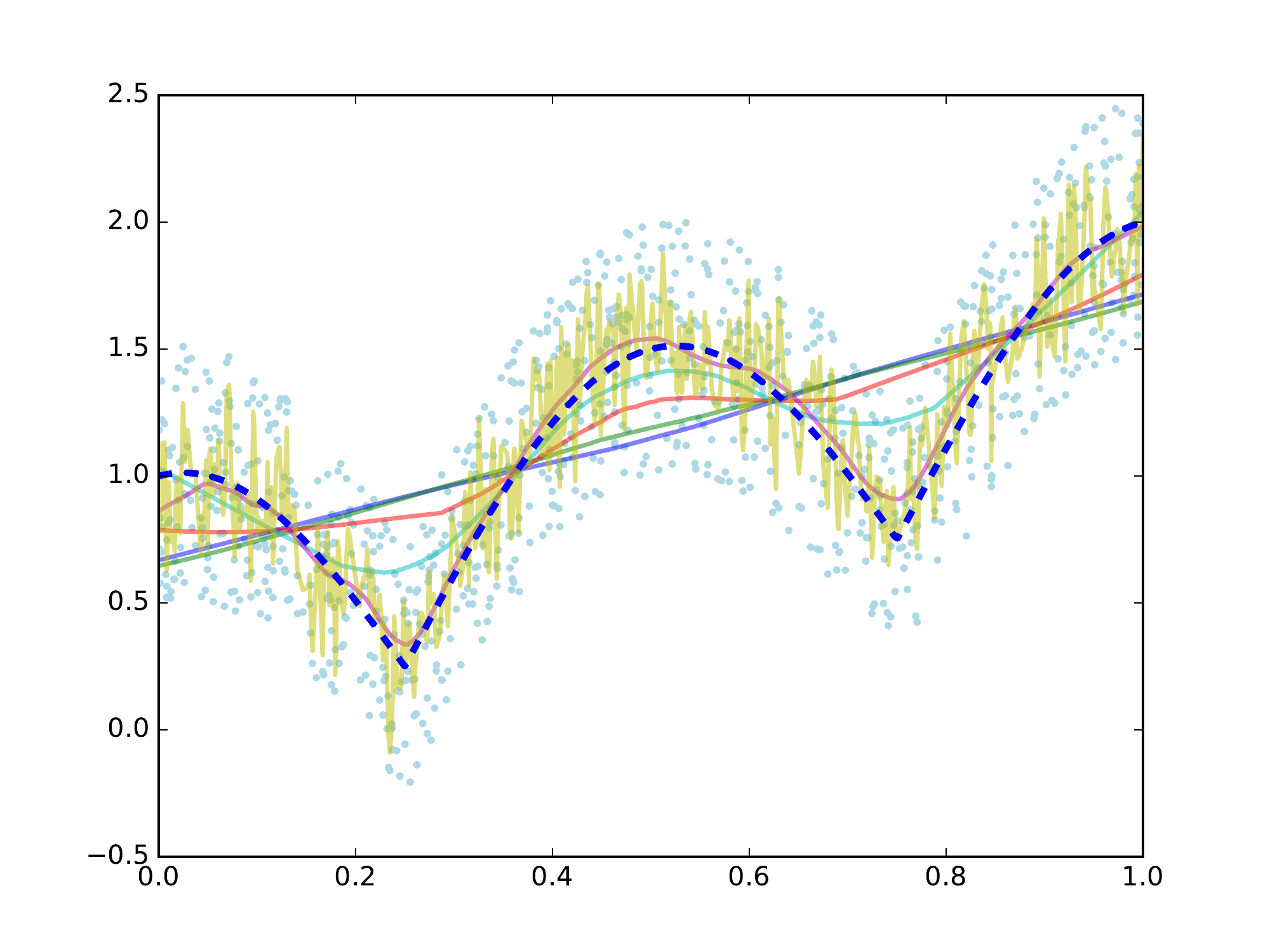}
\caption{Predictors (solid curves) generated by the Loess method for predicting  the test function (dashed curve) based on a noisy data set (blue dots) for various choices of the smoothness parameter. }
\label{fig:loess_test}
\end{figure}

As mentioned in the introduction most general nonparametric regression techniques' predictive performance hinges on the choice of a set of hyper-parameters. For instance, in the Loess method \cite{Cleveland1979} there is a smoothness parameter to be set a priori. For an illustration of the strong impact of this parameter to the predictor refer to Fig. \ref{fig:loess_test}.

We can see the behaviour ranges between over-smoothing and overfitting to the noise (e.g. the yellow curve). Unfortunately, for the Loess method we are not aware of a systematic method for finding the parameter in question. Of course, in principle, our approach we use for the KI regression approach might be applicable as well. That said, currently we are not sure how to determine a Lipschitz constant (or even gradient) of the test error as a function of the parameter that would yield an analogous approach to the POKI method we propose in this paper. Therefore, for the time being, we will restrict our comparisons to other methods where automatic (hyper-) parameter determination techniques are available.

\begin{figure*}
        \centering
        				  \subfigure[ Predictions.]{
    \includegraphics[width = 1\textwidth, height = .15\textheight]
								{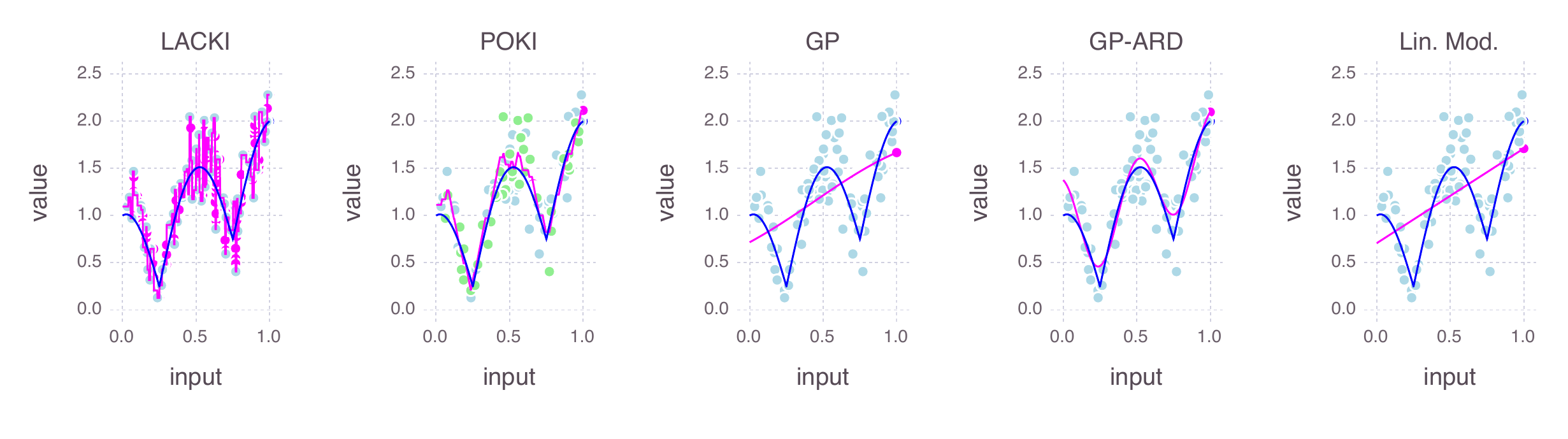}
    \label{fig:artiffunpred1d}
  } 	
  			   \caption{Exp.1. Predictions (magenta curve) of the target function $f: x\mapsto \abs{-\cos(2\pi \, x_1)}+x_1$ (blue curve) on one-dimensional input space with various models trained on a random sample of 84 data points for some of our learning methods. The training data is plotted as cyan dots; for POKI we have plotted the conditioning data in green.	   
Observe how the LACKI model overfits to the noisy data, while POKI manages to smooth out the noise and quite accurately predicts the ground-truth target function. Moreover, the GP with fixed hyper-parameters over-smoothes, while the GP with hyper-parameter optimisation (GP-ARD) manages to find hyper-parameters that lead to substantially better predictions.}\label{fig:test_artifdat1}
\end{figure*}

In particular, we trained predictors based on the following learning methods:

\begin{enumerate}
\item Linear regression model (\textbf{Lin. Mod.}) trained on the data with the least-squares method.
\item A Gaussian process (\textbf{GP}) provided with the correct observational noise level. The prior was based on a zero-mean function $\mu(\cdot) \equiv 0$ and a squared-exponential (SE) ARD-kernel $k(x,x') =  \omega \exp\bigl(- \sum_{i=1}^d \frac{(x_i-x_i)^2}{\ell_i}\bigr)$ with relevance length scale parameters $\ell_i$ chosen uniformly to be $\ell_i =0.5, \forall i$ and output-scale parameter chosen to be $\omega  =1$. 
\item A kinky inference predictor (\textbf{LACKI}) with $\metric(x,x'; \param) = \param \norm{x-x'}_\infty $ where the parameter was the Lipschitz constant $\param=L(n)$ lazily  estimated as per Eq. \ref{eq:lazyconstadaptdefnonoise}. This is equivalent to the LACKI method described in \cite{Calliess2016_LACKIarxiv} with hyper-parameter choices $\underline L =0$ and $\hestthresh =0$.

\item Our parameter optimised kinky inference approach (\textbf{POKI-LC}) where again $\metric(x,x'; \param) = \param \norm{x-x'}_\infty $.

\item Our parameter optimised kinky inference approach (\textbf{POKI-LC2}) trained as \textbf{POKI-LC} but where the parameter was optimised with Brent's method instead of Lipschitz optimisation.

\item Our parameter optimised kinky inference approach (\textbf{POKI-ARD}) where the metric was an ARD metric with relevance parameters optimised as described above.

\item A Gaussian process (\textbf{GP-ARD}) with a SE-ARD kernel as above, but whose relevance hyper-parameters were optimised with the conjugate-gradient approach to maximise the marginal log-likelihood \cite{GPbook:2006}.
%\item A Gaussian process (\textbf{GPopt1}) as above but with hyper-parameters optimised with the standard approach (cf.  \cite{GPbook:2006}) of utilising the BFGS-method to maximise the marginal log-likelihood of the data. The correct observational noise model was supplied.
%
%\item A Gaussian process trained as (\textbf{GPopt2}) as in GPopt1 but under the false assumption of too low observational noise level with standard deviation $0.01$.
\end{enumerate}
\textbf{Exp.1 :} As a first illustration, we considered regression on one-dimensional input space ($d=1$) based on a sample of 84 noise-corrupted data points. The results for a selection of the methods are depicted in Fig. \ref{fig:artiffunpred1d}.
We can see that the linear regression method accurately fitted the linear trend in the data but does not explain the nonlinear variation. For the given hyper-parameter settings, the GP was clearly over-smoothing the data. This might well have to do with the choice of assumed observational noise and length scale. To obtain better results we have also tried GP hyper-parameter optimisation on the data (GP-ARD). The conjugate gradient  optimiser managed to determine sensible hyper-parameters, thereby reducing the over-smoothing markedly. 
In contrast, the LACKI approach uncovered the nonlinearity well but overfitted to the noise in the data. Finally, our POKI-LC approach was able to smooth out the noise sufficiently well to offer a good fit to both the linear as well as to the non-linear trend in the target function. Being of comparable prediction accuracy as the solution found by GP-ARD, POKI's predictor was markedly less smooth than the GP's predictor with the chosen kernel and thereby, being able to fit the non-smooth parts of the target function better.\footnote{Of course, we could have handpicked a less smooth kernel class for the GP to potentially match the non-smooth parts of target better. However, we have opted for the SE class as this is the universal kernel class most commonly utilised in ``black-box'' learning.}

%We have also attempted hyper-parameter optimisation for training a GP (GPopt1 and GPopt2) on this data but found it difficult to produce reliably good outcomes (only about 1 in 10 times did the approach find good hyper-parameters that resulted in a posterior GP that fitted the data well). By contrast, POKI-LC reliably produced good predictions (for sample sizes greater than 30).

\begin{figure*}
        \centering
                  				  \subfigure[Abs. pred. error statistics.]{
    \includegraphics[width = 0.48\textwidth]
								{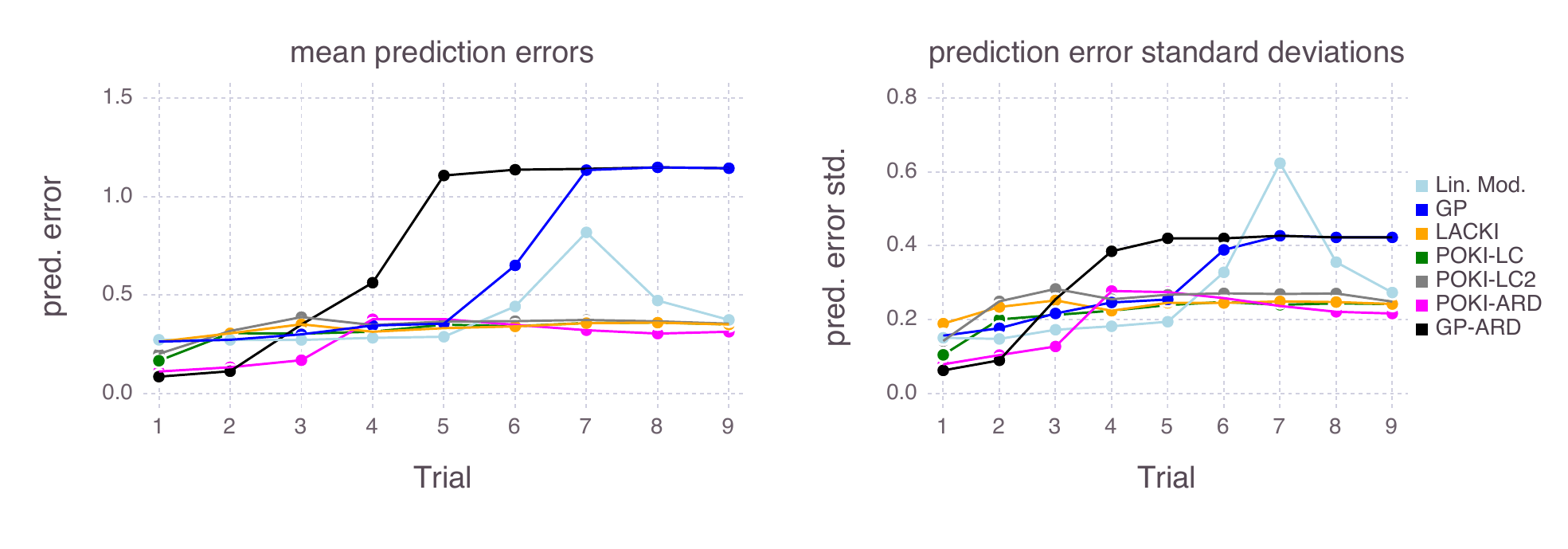}
    \label{fig:predtraj_vardim}
  } 	
        				  \subfigure[ Runtimes (log-sec.).]{
    \includegraphics[width = 0.48\textwidth]
								{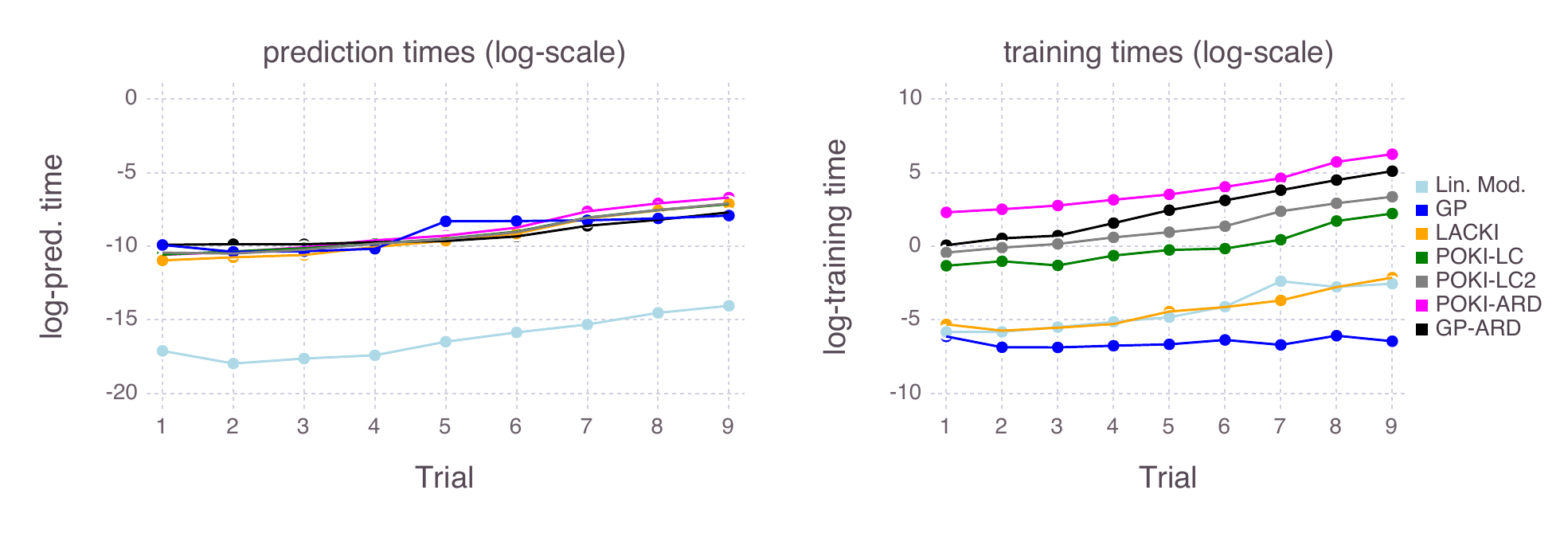}
    \label{fig:logTvardim}
  } 	
  			   \caption{Exp.2 (i). Fig. \ref{fig:predtraj_vardim}, left: Prediction error means ($\hat {\mathcal E_1}$) for the different trials. In trial $i$, the input space dimensionality was chosen to be $d=2^i$. The training data size was set to a fixed value of 150 training examples during all trials.  Fig. \ref{fig:logTvartex} depicts the logarithms of the pertaining records of runtimes (in seconds) for training the models (right) and the average prediction time for the test inputs (left). Not surprisingly, the training effort tends to increase with the number of parameters to be optimised. }\label{fig:test_artifdat1_vardim}
\end{figure*}

\begin{figure*}
        \centering
                  				  \subfigure[Abs. pred. error statistics.]{
    \includegraphics[width = 0.48\textwidth]
								{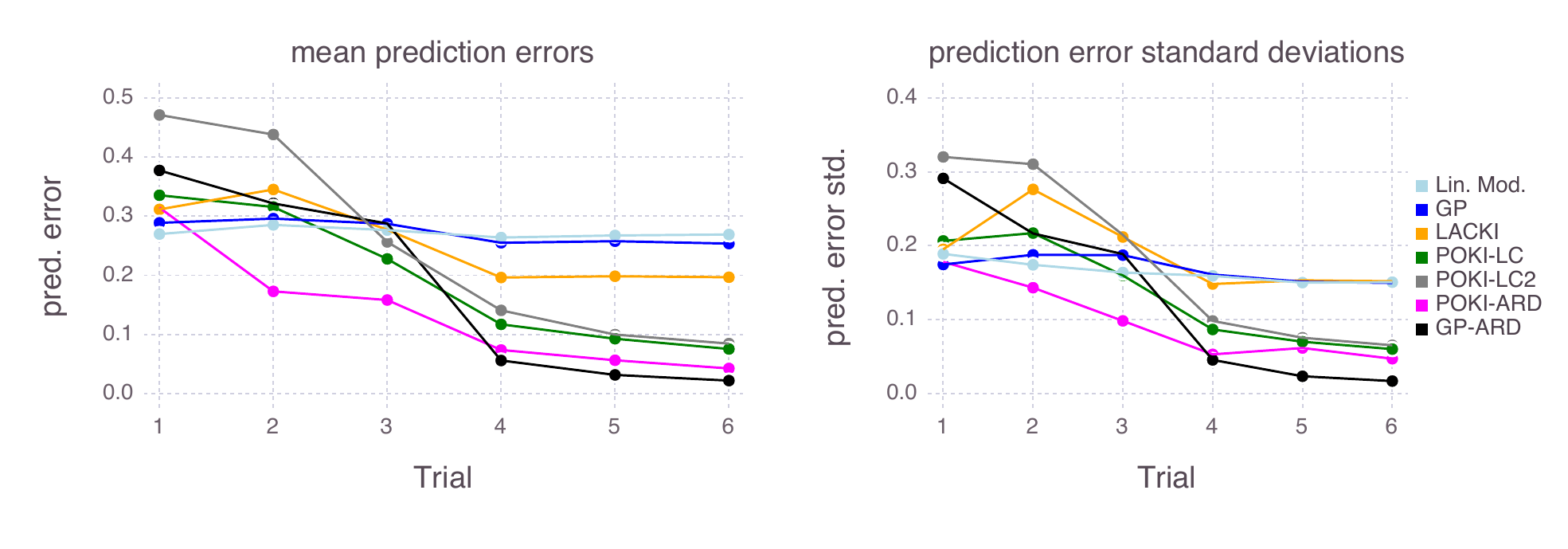}
    \label{fig:predtraj_vartex}
  } 	
        				  \subfigure[ Runtimes (log-sec.).]{
    \includegraphics[width = 0.48\textwidth]
								{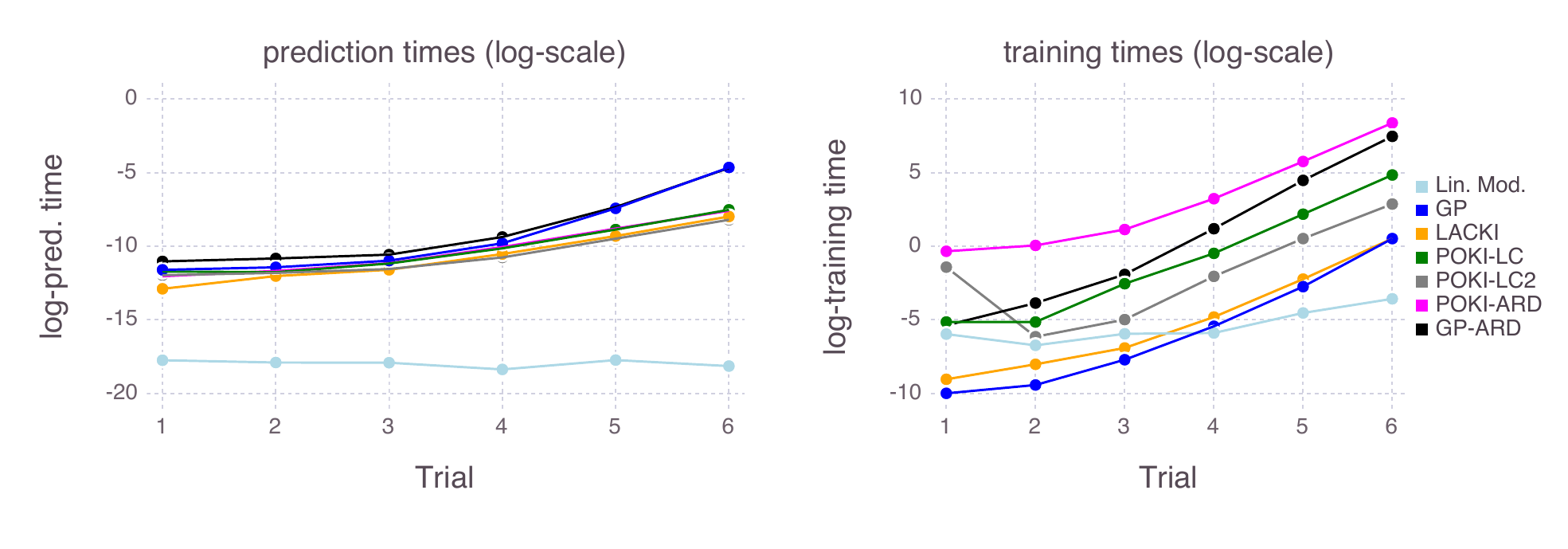}
    \label{fig:logTvartex}
  } 	
  			   \caption{Exp. 2 (ii). Fig. \ref{fig:predtraj_vartex} left: Prediction error means ($\hat {\mathcal E_1}$) for the different trials with increasing sample size. In all trials the input space dimensionality was chosen to be $d=2$. In trial $n$, the sample size $\abs{\data_n}$ was chosen to be $2^{d\,n}$. Fig. \ref{fig:logTvartex} depicts the logarithms of the pertaining records of the runtimes (in seconds) for training the models (right) and the average prediction time for the test inputs (left). Note how the prediction errors of most of the methods drops with increasing sample size. Successfully uncovering the low-dimensional structure, POKI-ARD tends to outperform or match the other methods in terms of $\hat {\mathcal E_1}$ error. }\label{fig:test_artifdat1_vartex}
\end{figure*}

\begin{figure*}
        \centering
                  				  \subfigure[Abs. pred. error statistics.]{
    \includegraphics[width = 0.48\textwidth]
								{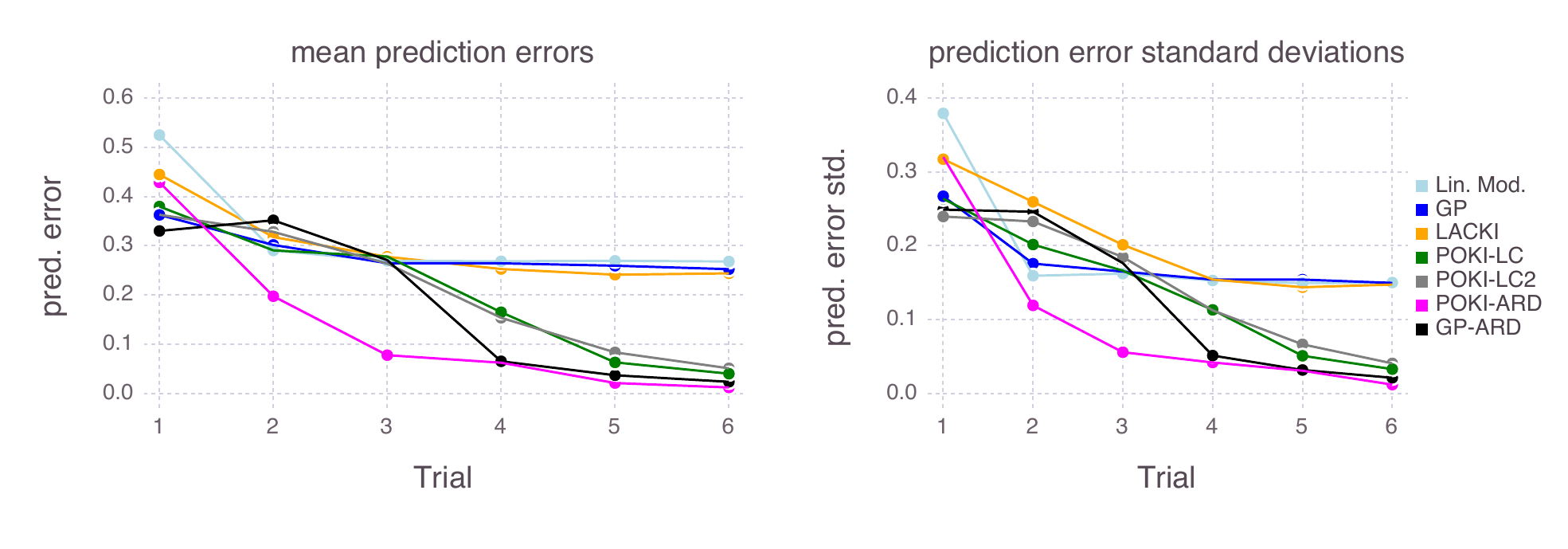}
    \label{fig:predtraj_vartex_bndnoise}
  } 	
        				  \subfigure[ Runtimes (log-sec.).]{
    \includegraphics[width = 0.48\textwidth]
								{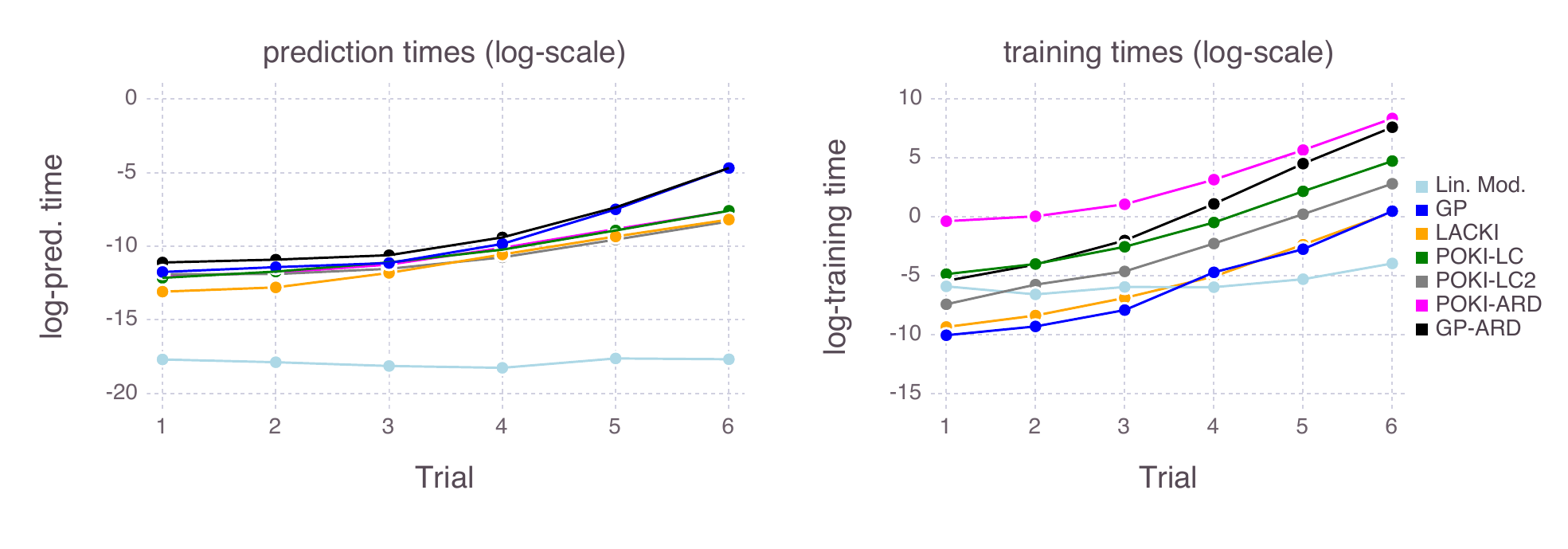}
    \label{fig:logTvartex_bndnoise}
  } 	
  			   \caption{Exp. 2 (iii). Repetition of Exp.2 (ii), but with uniformly distributed, bounded noise. Note how the prediction errors of most of the methods drops with increasing sample size. Successfully uncovering the low-dimensional structure, POKI-ARD tends to outperform or match the other methods in terms of $\hat {\mathcal E_1}$ error. }\label{fig:test_artifdat1_vartex_bndnoise}
\end{figure*}

\textbf{Exp.2 :} We desired to gain an impression of the performance of our POKI methods with (i) increasing input space dimensionality and (ii) in the limit of increasing sample size. In a third set of experiments (iii), we also the repeated the experiments in (ii) but for training data corrupted by noise $\nu$ drawn i.i.d. from a uniform over the intervall $[-0.5,0.5]$.
The results are depicted in Fig. \ref{fig:test_artifdat1_vardim}, Fig. \ref{fig:test_artifdat1_vartex} and Fig. \ref{fig:test_artifdat1_vartex_bndnoise}, respectively. As performance metrics of interest, we considered the quality of the prediction and the computational effort required for training as well as for predicting with the trained models. The predictive accuracy was estimated by the empirical absolute prediction error means (cf. Eq. \ref{eq:prederrest}) estimated on the basis of a test set sample of 4000 inputs drawn i.i.d. from a uniform over the input space. The test sample was drawn independently from the training examples. Runtime measurements were based on code written in Julia 4.7 running on a single core of a 2.5 GHz i7 processor on a laptop with 16GB RAM. Our evaluation of the GPs was based on the implementation provided by the \emph{GaussianProcesses.jl} library.

Examining the plots we note the following observations:
Firstly, for all POKI methods, the prediction error seemed to vanish with increasing sample size. 
Secondly, the ARD methods were able to yield better prediction accuracy. We attribute this to their ability to uncover the inherently one-dimensional structure of the functional relationship. In comparison to the tested alternatives, POKI-ARD seemed to be more reliable in finding good parameters (and hence, to yield superior prediction accuracy) than its competitors. We attribute this to the use of the global Lipschitz optimiser. 
This benefit comes at the cost of increased training duration in the current implementation of the optimiser in the ARD case, when the dimensionality of the data increases. Addressing this issue is deferred to future work.
In Exp.2 (ii) and (iii), it appeared that the Lipschitz optimisation-based hyper-parameter tuning seemed to be particularly beneficial for a small- to mid- range training data density (e.g. compare the prediction error plots of POKI-LC vs POKI-LC2 as well as POKI-ARD vs GP-ARD). Furthermore, in Exp.2 (i), it emerged that both GP-based predictors performed comparably poorly as the dimensionality of the input space grew (cf. Fig. \ref{fig:test_artifdat1_vardim}).

\subsection{Real data.}
We also tested the predictive power of our methods on two real data sets. 
With no access to the true target the performance estimates had to rely on the empirical sample prediction error (cf. Eq. \ref{eq:sampleprederr}). The first data set was 
the UCI-CCPP data set \cite{Tuefekci2014} where the task was to predict the output of a power plant based on sensor measurements and control inputs. The available data was partitioned randomly into a training and a test data set containing 10\% and 90\% of all available data, respectively.
The second data set was the Pumadyn data set \cite{Ghahramani1996}. Simulating the dynamics of the robot arm of a 560 Puma robot, the data has been   
used to showcase the potency of Gaussian processes for learning dynamical systems \cite{GPbook:2006}. The available data was partitioned randomly into a training and a test data set containing 60\% and 40\% of all available data, respectively.

The results are documented in Tab. \ref{tab:table2}. While the GP with hyperparameter optimisation (GP1) outperformed all other methods on the Pumadyn data set, POKI-ARD exhibits similarly low test set error mean. 
Furthermore, on the CCPP data the GP model performed very poorly, while our POKI methods were estimated to have the lowest test set error. 

\newcommand{\ra}[1]{\renewcommand{\arraystretch}{#1}}

\begin{table}
 %\begin{minipage}[h]{1.\textwidth}
%\begin{table}[h!]
  \centering
 % \caption{Sample statistics.}
 % \label{tab:table1}
  \begin{tabular}{ccccc}
    \toprule
    \emph{Data set} & \emph{Sample mean} & \emph{Sample std} & \emph{Size} & \emph{Input dim.}\\
    \midrule
    UCI-CCPP & 454.37 & 17.06& 9568 &4\\
    PumaDyn-8hn & 1.16 & 5.62& 4915 &8\\
    \bottomrule
  \end{tabular}
%\end{table}
%\end{minipage}
\caption{ Sample statistics.}
\label{tab:table1}
\end{table}

\begin{table}
%\begin{minipage}[h]{1.\textwidth}
\centering
 %\begin{table}[bht]   
 \centering
 \begin{footnotesize} 
 %\ra{1.3}
 \begin{tabular}{@{}lcccccc@{}}
 \toprule
\emph{Data set:}& \multicolumn{3}{c}{UCI-CCPP} & 
 \multicolumn{3}{c}{PumaDyn-8hn} \\
 \cmidrule(r){2-4} \cmidrule(l){5-7} 
\emph{Statistic:} & Mean& Std &    & Mean& Std & Median \\
 \midrule
 
 Lin. Regr.	& $3.64$ & $2.81$ & 	& 3.68 	& 2.49 & 3.39 \\
GP1 & $206.18$	&$144.98  $&  	& \textbf{2.63}	& 2.15 & \textbf{2.14} \\
LACKI & $3.40$	& $3.44$ &  	& 3.28 	& 2.55 & 2.69 \\
POKI-LC & $3.25 $	&$ 2.98$ & 	& 3.25 	& 2.43 &  2.76\\
POKI-ARD & $\textbf{2.60}$& $2.79$ & & 2.81	& 2.07	& 2.44\\
%% %$dir=0$\\
%% %$c$ & 0.0357& 1.2473& 0.2119&& 0.3593& -0.2755& 2.1764\\
%% %$c$ & -17.9048& -37.1111& 8.8591&& -30.7381& -9.5952& -3.0000\\
%% %$c$ & 105.5518& 232.1160& -94.7351&& 100.2497& 141.2778& -259.7326\\
 \bottomrule
 \end{tabular}
 \end{footnotesize}
%
 %\caption{Measured abs. test error statistics. }
 %\label{Tab:perf}
 %\end{table}
% \end{minipage}
%
\caption{Measured abs. test error statistics. }
\label{tab:table2}
\end{table}

\vspace{-1em}

\subsection{Conclusions}
We developed POKI, a method for nonparametric regression that optimises parameters of a chosen pseudo-metric with the aim to maximise predicitve performance. Our approach can be applied to automated determination of Lipschitz constants in nonlinear set membership methods thereby addressing an open problem in data-driven control in a more principled manner than previously proposed methods \cite{Milanese2004,Canale2014}. %By optimising for predictive performance it offers a more principled, goal orientated alternative to previously proposed methods that employed Lipschitz constants of estimated via parametric proxy models fitted to the data. 
In addition, we have also given examples where the automatically learned (hyper-) parameters 
were utilised for automated relevance determination and detected and leveraged periodicity in the data.

We have found that our POKI methods work. They can compensate for noise in the data without the necessity to know the correct noise distributions. They are flexible to learn nonlinear, non-smooth functions. 
While the performance will not always be better than competing methods (such as GPs), an advantage of our approach is that the inference only involves basic computational steps that we would expect to be numerically robust and executable even on reduced instruction set micro-controllers. Furthermore, the particular mathematical properties of the Lipschitz interpolation rule allow us to easily determine Lipschitz constants of the empirical prediction errors. This in turn allows us to determine hyper-parameters by global Lipschitz optimisation, avoiding the pitfalls of local optima. Therefore, our approach can be more flexible and reliable to perform well in a large variety of problem domains and data sets without any manual tweaking. This is in contrast to many other methods (such as GPs or deep learning methods), where the predictive performance can be sensitive to a priori choices, e.g. of priors, hyper-parameters, optimisers and initialisations.

With the present choice of Lipschitz optimiser, problems involving high-dimensional hyper-parameters can render the determination of the approximately optimal hyper-parameter computationally intractable. Future work intends to explore the utilisation or development of alternative Lipschitz optimisation approaches that promise improved scalability (for a fixed error bound) in the dimensionality of the optimisation problem (e.g. \cite{Zhang1995}).

We observed that our approach can smooth out noise and hence, can yield accurate prediction of the ground truth even in the presence of additive observational noise.
A theoretical challenge that remains to be investigated is the asymptotics of the true $\mathcal L_1$-prediction error in the limit of increasing number of noisy data points.
We would hope that these results would allow us to derive probabilistic guarantees for a data-driven controller that combines our learning method with stochastic MPC approaches. This would provide an important extension to existing work of NSM-based MPC that had to rely on the knowledge of the Lipschitz constant \cite{Canale2014,calliess2014_thesis}.

Finally, we would like to point out that Lipschitz interpolation rules (i.e. KI or NSM predictors) have a mathematical structure that made it particularly easy to calculate a Lipschitz bound on the empirical $\ell_1$ prediction loss, facilitating global Lipschitz optimisation of the hyper-parameters. However, it might be worthwhile exploring in how far the Lipschitz optimisation approach can also be successfully employed in automated (hyper-) parameter tuning of other machine learning methods.

\section{Acknowledgements}
I would like to thank Carl Rasmussen and Jan Maciejowski for helpful discussions and encouraging feedback. Also, funds via EPSRC NMZR/031 RG64733 are gratefully acknowledged.
%\newpage
\bibliographystyle{plain}
\bibliography{lit}

\appendix 
\subsection{H\"older arithmetic and optimisation }
This work utilised global Lipschitz optimisation and, to facilitate this, we derived Lipschitz constants for various functions pointing to the algorithm and results stated in 
\cite{calliess2014_thesis}. For convenience, we restate these results at this point followed by an outline of a simple method for bounded H\"older optimisation we utilised in this work. 

\subsubsection{H\"older arithmetic}

As a slight generalisation of Lipschitz continuity, we say that a function $f: \inspace \to \outspace$ is H\"older continuous on $I \subseteq \inspace$ with (H\"older) constant $L$ and (H\"older) exponent $p$ if $\metric_\outspace(f(x),f(x')) \leq L \metric_\inspace(x,x') \forall x,x' \in I$.

\begin{lem}[H\"older arithmetic] \label{lem:Hoeldarithmetic}
Let, $I,J \subset \inspace$ where $\inspace$ is a metric space endowed with metric $\metric$. Let $f : \inspace \to \Real$ be H\"older on $I$ with constant $L_I (f) \in \Real_+$ 
and $g :\inspace \to \Real$ be H\"older on $J$ with constant $L_J (g) \in \Real_+$. Assume both functions have the same H\"older exponent $p \in (0,1]$. That is, $\forall x, x' \in \inspace: \abs{f(x)-f(x')} \leq L(f) \metric(x,x')^p$ and  $\forall x, x' \in \inspace: \abs{g(x)-g(x')} \leq L(g)  \metric(x,x')^p$.
We have:

\begin{enumerate}
	\item Mapping $x \mapsto |f(x)|$ is H\"older on $I$ with constant $L_I(f)$ and exponent $p_f$.
	\item If $g$ is H\"older on all of $J=f(I)$ the concatenation $g \circ f: t \mapsto g(f(t))$ is H\"older on $I$ with constant 
	      $L_I(g \circ f) \leq$ $L_J (g) \, L_I^p(f)$ and exponent $p^2$.
	\item Let $r \in \Real$. $r \, f: x \mapsto r \, f(x)$ is H\"older on $I$ having a constant $L_I (r \,f) = |r| \, L_I(f)$.
	\item $f+g: x \mapsto f(x) + g(x)$ has H\"older constant at most $L_I(f) + L_J(g)$.
	\item Let $m_f = \sup_{x\in \inspace } f(x)$ and $m_g = \sup_{x \in \inspace } g(x)$. Product function $f\cdot g: x \mapsto f(x) \, g(x)$ has H\"older exponent $p$ and a H\"older constant on $I$ which is at most $(m_f \, L_J(g)+ m_g \, L_I(f))$.
	%\item Let $\tilde h(x) = \min\{f(x),g(x) \}$, $h(x) = \max\{f(x), g(x) \}, \forall x \in \inspace  \cap J$. We have $L_{I \cap J}(h) \leq \max\{L_I(f),L_J(g)\}$ and $L_{I \cap J}(\tilde h) \leq \max\{L_I(f),L_J(g)\}$.	
	\item For some countable index set $\indsett$, let the sequence of functions $f_i$ be H\"older with exponent $p$ and constant $L(f_i)$ each. Furthermore, let $H(x) =\sup_{i \in \indsett} f_i(x) $ and $h(x) := \inf_{i \in \indsett} f_i(x)$ be finite for all $x$. Then $H,h$ are also H\"older with exponent $p$ and have a H\"older constant which is at most $\sup_{i \in \indsett} L(f_i)$.
	\item Let $b := \inf_{x \in \inspace }| f(x)| > 0$ and let 
	$\phi(x) = \frac{1}{f(x)}, \forall x \in \inspace$ be well-defined.  
	      Then $L_I(\phi) \leq b^{-2} \, L_I(f)$.  
	\item Let $p=1$ (that is we consider the Lipschitz case), let $I$ be convex and $\metric(x,x') = \norm{x-x'}$ where $\norm{\cdot}$ is a norm that induces a sub-multiplicative matrix norm (e.g. all $p-$ norms are valid). $f$ cont. differentiable on I $\Rightarrow$ $L_I(f) \leq \sup_{x \in I } \norm{\nabla f(x)}. $ 
	For one-dimensional input space, $\inspace = \Real$, $L_I(f) = \sup_{x \in I } \abs{\nabla f(x)}$ is the smallest Lipschitz number. 
	 \item Let $c \in \Real$, $f( t) = c, \forall x \in I $. Then $f$ is H\"older continuous with constant $L_I(f) =0$ and for any coefficient $p_f \in \Real$.  
	\item $L_I(f^2) \leq 2 \, L_I(f)\, \sup_{t \in I} f\,$.
	\item With conditions as in 8), and input space dimension one, we have $\forall q \in \mathbb Q : L_I(f^q) = |q| \,\sup_{\tau \in \indset } |f^{q-1}(\tau) \, \dot f(\tau)| $.
\end{enumerate}
\end{lem} 
\begin{proof}
Refer to \cite{calliess2014_thesis}.
\end{proof}

\subsubsection{H\"older continuity for optimisation with bounds}
%\jcom{Nat suggests to cite no regret optim paper}
%\subsection{A simple bounded optimisation procedure for Hoelder continuous functions on multi-dimensional domains} 
\label{sec:Hoelder_opt_basic_brief}
In the presence of a known Lipschitz constant, optimisation of a Lipschitz continuous function can be done with Shubert's algorithm \cite{Shubert:72}. Of importance to us at various parts of the thesis is, that it also provides non-asymptotic error bounds.  Unfortunately, it is limited to minimising functions on one-dimensional input domains. 
For multi-dimensional domains DIRECT \cite{direct:93} is a global optimiser that is popular, not least because no Lipschitz constant is required. On the flip side, it does not provide the error bounds afforded by Shubert's method. Furthermore, we will sometimes be interested in bounds on the maxima and minima for the more general case, where the functions to be optimised are H\"older continuous with respect to normed spaces.  
%This is fine when we attempt to find $\bar v$ for a GP. However, for random fields, i.e. multi-input GPs, we require an optimisation algorithm capable of finding a conservative upper bound on the maximum of the kernels $k(\cdot,x_i)$ even for vector inputs. Fortunately, a naive method one could use, even for H\"older-continuous functions, is as follows:
We are currently working on such a method, as well as on a corresponding cubature algorithm. 
As it is work in progress and its exposition is beyond the scope of this work, we will merely sketch a naive version that provides some desired bounded optimisation for H\"older continuous functions:

For simplicity, assume the input and output space metrics are induced by the maximum norms. For example, $\metric_\inspace(x,x') = \norm{x-x'}_\infty$.\footnote{Due to norm-equivalence relationships we can bound each norm by each other norm by multiplication with a well-known constant (see e.g. \cite{koenigsberger:2000}). Therefore, knowing a H\"older-constant with respect to one norm immediately yields a constant with respect to another, e.g. $\norm{\cdot}_\infty \leq \norm{\cdot}_2 \leq \sqrt{d} \norm{\cdot}_\infty$ where $d$ is the dimension of the input space.} 
Let $f: I \subset \Real^d \to \Real$ be a H\"older continuous function with H\"older constant $L$ and exponent $p$. Let  
$S = \{(s_i, f_i) | i =1,...,N\} $ be a sample of target function $f$ with $f_i = f(s_i)$, $s_i \in I, \forall i$. 

Assume $I$ is a hyper-rectangle and  let  $J_1,...,J_N \subset I$ be a partition of the domain $I$ such that each sub-domain $J_j$ is a hyperrectangle containing $s_j$.
(More generally, to avoid $I$ having to be a hyperrectangle itself, one could alternatively assume that the $J_1,\ldots,J_N$ are a covering of $I$. That is, $I \subset \cup_i J_i$.)
 
By H\"older continuity $\forall x \in J_i: \abs{f_i - f(x)} \leq L \norm{x - s_i}_\infty^p$for some $p \in (0,1], L \geq 0$. 
Let  $a_1,\ldots,a_d >0$, $b_1,\ldots,b_d>0$ be defined such that $J_i =  s_i + ([-a_1,b_1] \times...\times [-a_d,b_d])$ where the addition is defined point-wise. Furthermore, let 
\begin{equation}
	D_j:= \max\{a_1,\ldots,a_d, b_1,\ldots,b_d\}
\end{equation}
 be the maximal distance of $s_i$ to the boundary of $J_j$.

Then, $\forall x \in J_j: f(x)  \leq f_j + L \max_{x \in J_j} \norm{x - s_j}_\infty^p =f_j + L D_j^p =: q_i$. 

Since, the hyperrectangles formed a covering of domain $i$, the desired upper bound on the maximum is $\max_{j=1}^N q_j \geq \max_{x \in I} f(x)$.
By the same argument, a lower bound on the maximum based on the finite sample can be obtained as:
$\min_{j=1}^N f_j - L D_j^p  \leq \min_{x \in I} f(x)$.

In conclusion, we have 
\begin{align}
\underline M := \min_{j=1\,...,N} f_j - L \, D_j^p  \leq \min_{x \in I} f(x) \label{eq:minhoeldineqfinsample}\\
\overline M := \max_{j=1,...,N} f_j + L \, D_j^p  \geq \max_{x \in I} f(x). \label{eq:maxhoeldineqfinsample}
\end{align}
A \textit{batch} algorithm would get a sample, construct a partition $J_1,...,J_N$ for that sample and then compute the minimum and maximum bounds $\underline M, \overline M$ given above.

For our tests in this work, we have implemented an \textit{adaptive} algorithm for bounded optimisation which incrementally expands the given sample. That is, it incrementally partitions domain $I$ into increasing small sub-hyperrectangles until the computational budget is expended or the bounds on the maxima and minima have shrunk satisfactorily. 

The exploration criterion of where to obtain the next function sample is tailored to finding $\overline M$ or  $\underline M$, depending on the task at hand.

For instance, in a minimisation problem we chose to obtain a new sample point in the hyperrectangle $J_{j_*}$ where  $ j_* = \arg\min_{j=1\,...,N} f_j - L \, D_j^p $ is the current rectangle where our present bounds allow for the lowest value to be.

%based on reduction of the error $E=\overline M - \underline M$. That is, we desire to add a sample such that $E$ is minimal after the new sample is inserted. 

%
%As an illustration, consider the following situation: We are given the covariance function
%$k(t,t') := \exp(- \abs{t-t'})$. One a priori upper bound on $\bar v$ would be to choose $k(t,t) = 1$.
%If we have training data $\data$, we can compute $\bar v$ as per  Eq. \ref{eq:barvQ}.
%
%To this end, we notice that $k(\cdot,t') = f \circ g (\cdot)$ with $f(t) = \exp(- t)$ and $g(t) = \abs{t-t'}$. Both functions are Lipschitz. That is, they are H\"older with exponent $p_f=p_g = 1$ and constants $L_f=L_g = 1$.

\end{document}